# Finding Approximate POMDP Solutions Through Belief Compression

**Nicholas Roy** NICKROY@MIT.EDU
*Massachusetts Institute of Technology,*
*Computer Science and Artificial Intelligence Laboratory*
*Cambridge, MA*

**Geoffrey Gordon** GGORDON@CS.CMU.EDU
*Carnegie Mellon University, School of Computer Science*
*Pittsburgh, PA*

**Sebastian Thrun** THRUN@STANFORD.EDU
*Stanford University, Computer Science Department*
*Stanford, CA*

## Abstract

Standard value function approaches to finding policies for Partially Observable Markov Decision Processes (POMDPs) are generally considered to be intractable for large models. The intractability of these algorithms is to a large extent a consequence of computing an exact, optimal policy over the entire belief space. However, in real-world POMDP problems, computing the optimal policy for the full belief space is often unnecessary for good control even for problems with complicated policy classes. The beliefs experienced by the controller often lie near a structured, low-dimensional subspace embedded in the high-dimensional belief space. Finding a good approximation to the optimal value function for only this subspace can be much easier than computing the full value function.

We introduce a new method for solving large-scale POMDPs by reducing the dimensionality of the belief space. We use Exponential family Principal Components Analysis (Collins, Dasgupta, & Schapire, 2002) to represent sparse, high-dimensional belief spaces using small sets of learned features of the belief state. We then plan only in terms of the low-dimensional belief features. By planning in this low-dimensional space, we can find policies for POMDP models that are orders of magnitude larger than models that can be handled by conventional techniques.

We demonstrate the use of this algorithm on a synthetic problem and on mobile robot navigation tasks.

## 1. Introduction

Decision making is one of the central problems of artificial intelligence and robotics. Most robots are deployed into the world to accomplish specific tasks, but the real world is a difficult place in which to act—actions can have serious consequences. Figure 1(a) depicts a mobile robot, Pearl, designed to operate in the environment shown in Figure 1(b), the Longwood retirement facility in Pittsburgh. Real world environments such as Longwood are characterized by uncertainty; sensors such as cameras and range finders are noisy and the entire world is not always observable. A large number of state estimation techniques explicitly recognize the impossibility of correctly identifying the true state of the world (Gutmann, Burgard, Fox, & Konolige, 1998; Olson, 2000; Gutmann & Fox, 2002; Kanazawa,

Koller, & Russell, 1995; Isard & Blake, 1998) by using probabilistic techniques to track the location of the robot. Such state estimators as the Kalman filter (Leonard & Durrant-Whyte, 1991) or Markov localization (Fox, Burgard, & Thrun, 1999; Thrun, Fox, Burgard, & Dellaert, 2000) provide a (possibly factored, Boyen & Koller, 1998) distribution over possible states of the world instead of a single (possibly incorrect) state estimate.

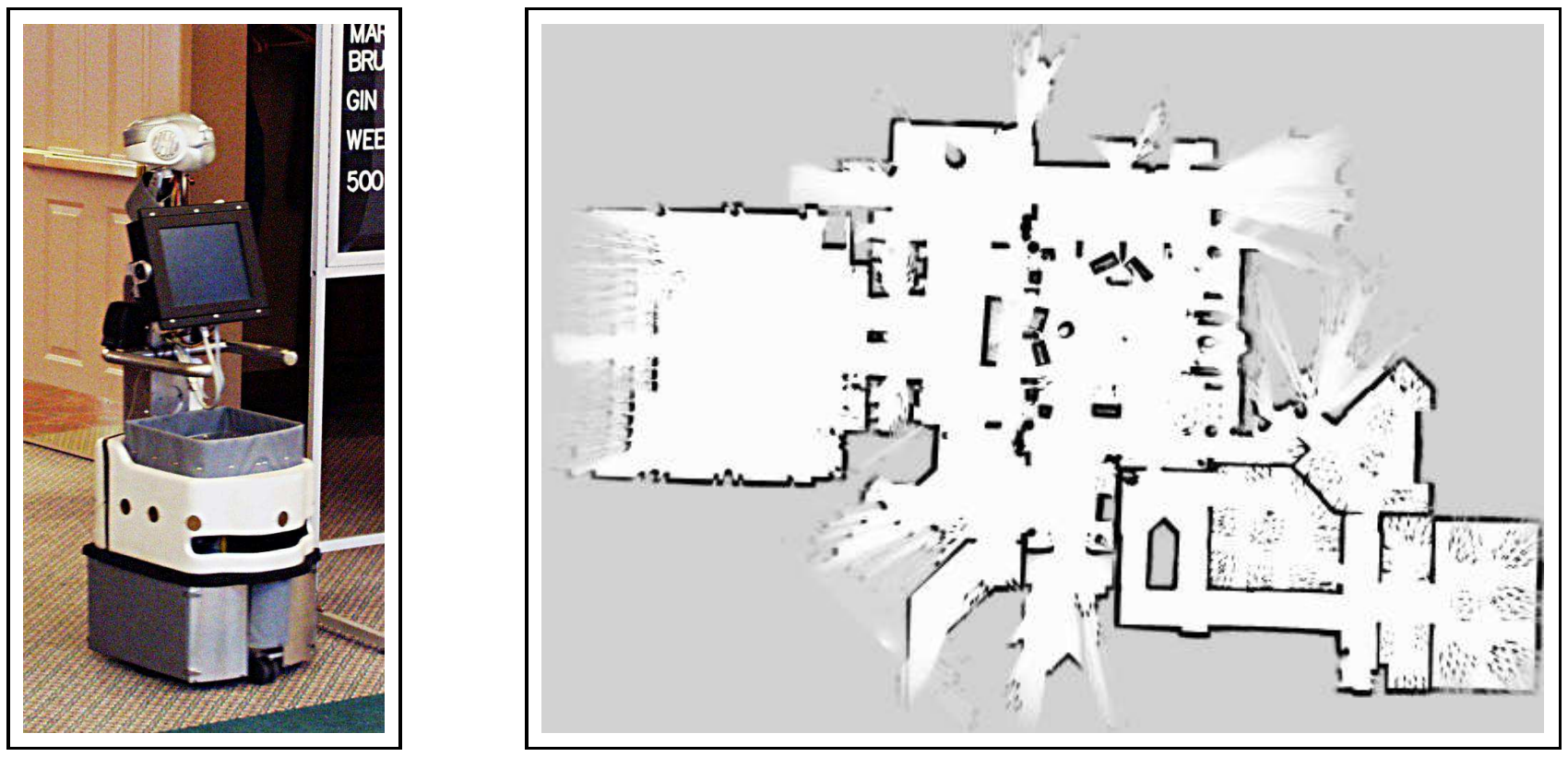

(a) (b)

**Figure 1**: A planner for the mobile robot Pearl, shown in (a), must be able to navigate reliably in such real environments as the Longwood at Oakmont retirement facility, shown in (b). The white areas of the map are free space, the black pixels are obstacles, and the grey areas again are regions of map uncertainty. Notice the large open spaces, and many symmetries that can lead to ambiguity in the robot's position. The map is $53.6m \times 37.9m$, with a resolution of $0.1m \times 0.1m$ per pixel.

In contrast, controllers such as motion planners, dialogue systems, etc. rarely model the same notions of uncertainty. If the state estimate is a full probability distribution, then the controller often uses a heuristic to extract a single "best" state, such as the distribution's mean or mode. Some planners compensate for the inevitable estimation errors through robust control (Chen, 2000; Bagnell & Schneider, 2001), but few deployed systems incorporate a full probabilistic state estimate into planning. Although the most-likely-state method is simple and has been used successfully by some real applications (Nourbakhsh, Powers, & Birchfield, 1995), substantial control errors can result when the distribution over possible states is very uncertain. If the single state estimate is wrong, the planner is likely to choose an unreasonable action.

Figure 2 illustrates the difference between conventional controllers and those that model uncertainty. In this figure, the robot must navigate from the bottom right corner to the top left, but has limited range sensing (up to $2m$) and noisy dead reckoning.[1] The impoverished

1. For the purposes of this example the sensing and dead reckoning were artificially poor, but the same phenomenon would occur naturally in larger-scale environments.

sensor data can cause the robot's state estimate to become quite uncertain if it strays too far from environmental structures that it can use to localize itself. On the left (Figure 2a) is an example trajectory from a motion planner that has no knowledge of the uncertainty in the state estimate and no mechanism for taking this uncertainty into account. The robot's trajectory diverges from the desired path, and the robot incorrectly believes it has arrived at the goal. Not shown here are the state estimates that reflect the high uncertainty in the robot position. On the right (Figure 2b) is an example trajectory from a controller that can model the positional uncertainty, take action to keep the uncertainty small by following the walls, and arrive reliably at the goal.

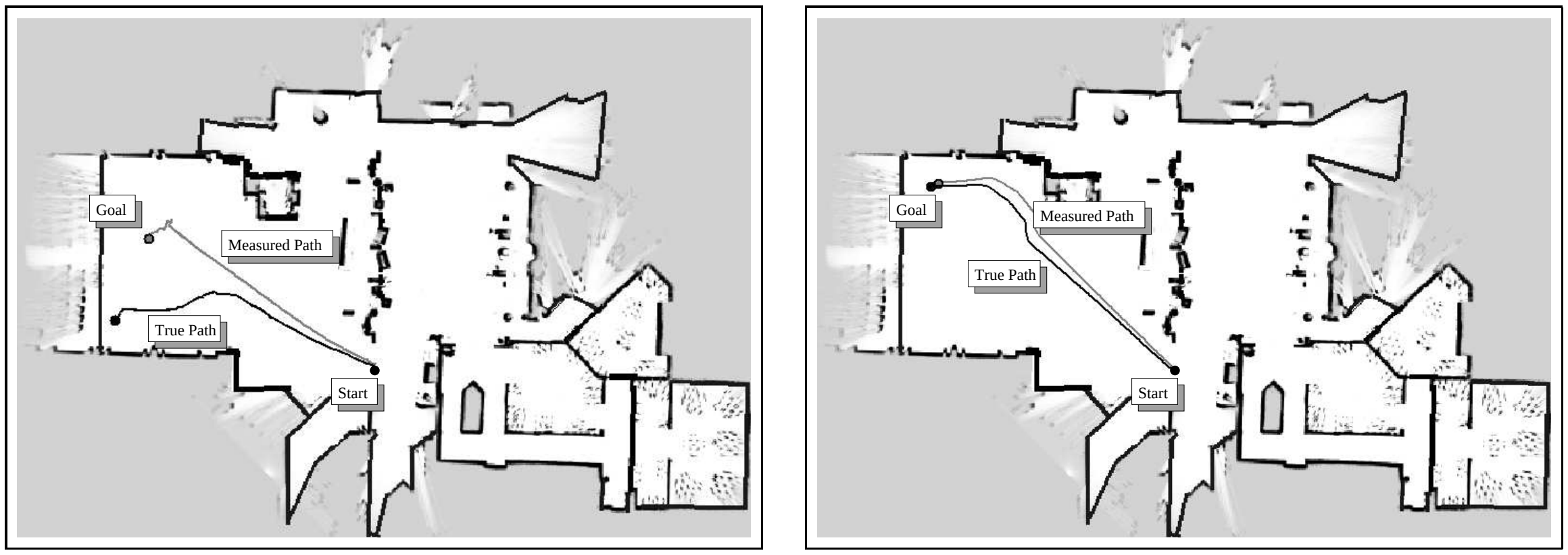

(a) Conventional controller

(b) Robust controller

**Figure 2**: Two possible trajectories for navigation in the Longwood at Oakmont environment. The robot has limited range sensing (up to $2m$) and poor dead-reckoning from odometry. (a) The trajectory from a conventional motion planner that uses a single state estimate, and minimizes travel distance. (b) The trajectory from a more robust controller that models the state uncertainty to minimize travel distance and uncertainty.

The controller in Figure 2(b) was derived from a representation called the partially observable Markov decision process (POMDP). POMDPs are a technique for making decisions based on probabilistic estimates of the state of the world, rather than on absolute knowledge of the true state. A POMDP uses an *a priori* model of the world together with the history of actions taken and observations received in order to infer a probability distribution, or "belief", over the possible states of the world. The controller chooses actions, based upon the current belief, to maximize the reward it expects to receive over time.

The advantage to using POMDPs for decision making is that the resulting policies handle uncertainty well. The POMDP planning process can take advantage of actions that implicitly reduce uncertainty, even if the problem specification (e.g., the reward function) does not explicitly reward such actions. The disadvantage to POMDPs is that finding the optimal policy is computationally intractable. Existing techniques for finding exact optimal plans for POMDPs typically cannot handle problems with more than a few hundred states (Hauskrecht, 2000; Zhang & Zhang, 2001). Most planning problems involving real, physical systems cannot be expressed so compactly; we would like to deploy robots that plan over thousands of possible states of the world (e.g., map grid cells), with thousands of possible observations (e.g., laser range measurements) and actions (e.g., velocities).

In this paper, we will describe an algorithm for finding approximate solutions to real-world POMDPs. The algorithm arises from the insight that exact POMDP policies use unnecessarily complex, high-dimensional representations of the beliefs that the controller can expect to experience. By finding low-dimensional representations, the planning process becomes much more tractable.

We will first describe how to find low-dimensional representations of beliefs for real-world POMDPs; we will use a variant of a common dimensionality-reduction technique called Principal Components Analysis. The particular variant we use modifies the loss function of PCA in order to better model the data as probability distributions. Using these low-dimensional representations, we will describe how to plan in the low-dimensional space, and conclude with experimental results on robot control tasks.

## 2. Partially Observable Markov Decision Processes

A partially observable Markov decision process (POMDP) is a model for deciding how to act in "an accessible, stochastic environment with a known transition model" (Russell and Norvig (1995), pg. 500). A POMDP is described by the following:

- a set of states $\mathcal{S} = \{s_1, s_2, \dots s_{|\mathcal{S}|}\}$
- a set of actions $\mathcal{A} = \{a_1, a_2, \dots, a_{|\mathcal{A}|}\}$
- a set of observations $\mathcal{Z} = \{z_1, z_2, \dots, z_{|\mathcal{Z}|}\}$
- a set of transition probabilities $T(s_i, a, s_j) = p(s_j|s_i, a)$
- a set of observation probabilities $O(z_i, a, s_j) = p(z_i|s_j, a)$
- a set of rewards $R : \mathcal{S} \times \mathcal{A} \mapsto \mathbb{R}$
- a discount factor $\gamma \in [0, 1]$
- an initial belief $p_0(s)$

The transition probabilities describe how the state evolves with actions, and also represent the Markov assumption: the next state depends only on the current (unobservable) state and action and is independent of the preceding (unobserved) states and actions. The reward function describes the objective of the control, and the discount factor is used to ensure reasonable behaviour in the face of unlimited time. An optimal policy is known to always exist in the discounted ($\gamma < 1$) case with bounded immediate reward (Howard, 1960).

POMDP policies are often computed using a value function over the belief space. The value function $V_\pi(b)$ for a given policy $\pi$ is defined as the long-term expected reward the controller will receive starting at belief $b$ and executing the policy $\pi$ up to some horizon time, which may be infinite. The optimal POMDP policy maximizes this value function. The value function for a POMDP policy under a finite horizon can be described using a piece-wise linear function over the space of beliefs. Many algorithms compute the value function iteratively, evaluating and refining the current value function estimate until no further refinements can improve the expected reward of the policy from any belief. Figure 3(a) shows the belief space for a three-state problem. The belief space is the two-dimensional, shaded simplex. Each point on the simplex corresponds to a particular belief (a three-dimensional vector), and the corners of the simplex represent beliefs where the state is known with 100% certainty. The value function shown in Figure 3(b) gives the long-term expected reward of a policy, starting at any belief in the simplex.

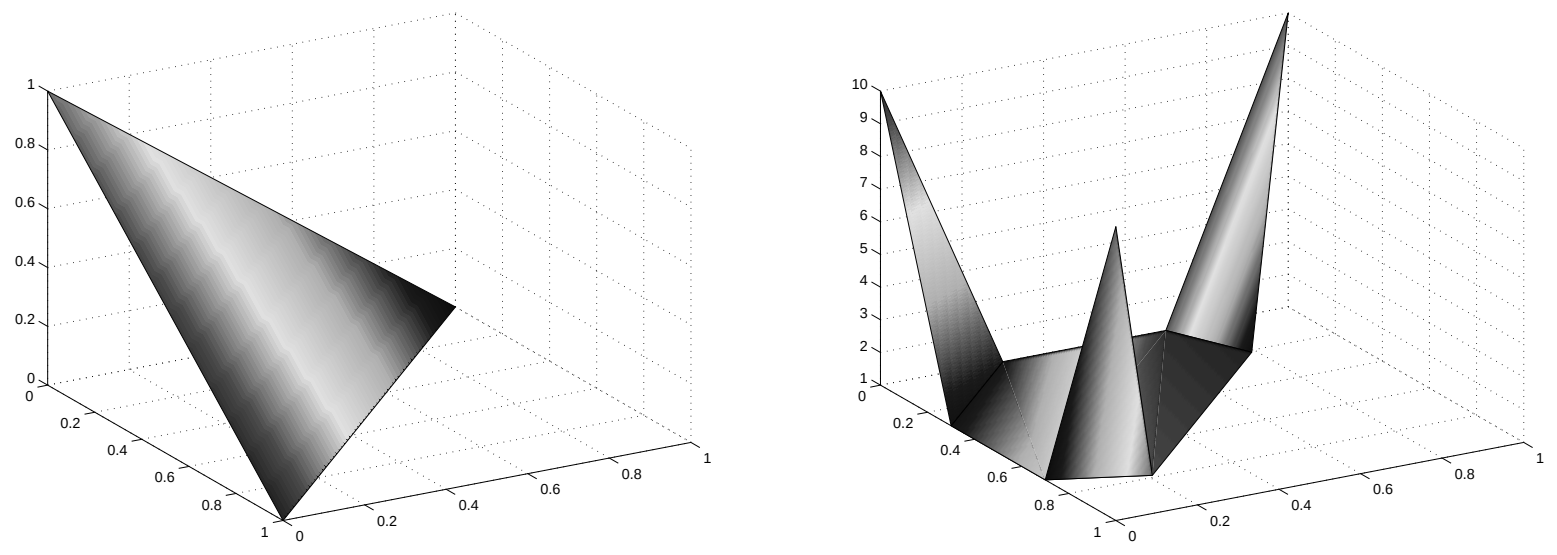

(a) The belief space (b) The value function

**Figure 3**: (a) The belief space for a three-state problem is the two-dimensional, shaded simplex. (b) A value function defined over the belief space. For the purposes of visualization, the set of beliefs that constitutes the belief space shown in (a) has been projected onto the $XY$ plane in (b); the value function then rises along the positive $Z$ axis. Each point in the belief space corresponds to a specific distribution, and the value function at that point gives the expected reward of the policy starting from this belief. The belief space (and therefore the value function) will have one fewer dimension than the total number of states in the problem.

The process of evaluating and refining the value function is at the core of why solving POMDPs is considered to be intractable. The value function is defined over the space of beliefs, which is continuous and high-dimensional; the belief space will have one fewer dimension than the number of states in the model. For a navigation problem in a map of thousands of possible states, computing the value function is an optimization problem over a continuous space with many thousands of dimensions, which is not feasible with existing algorithms.

However, careful consideration of some real-world problems suggests a possible approach for finding *approximate* value functions. If we examine the beliefs that a navigating mobile robot encounters, these beliefs share common attributes. The beliefs typically have a very small number of modes, and the particular shape of the modes is fairly generic. The modes move about and change in variance, but the ways in which the modes change is relatively constrained. In fact, even for real world navigation problems with very large belief spaces, the beliefs have very few degrees of freedom.

Figure 4(a) illustrates this idea: it shows a typical belief that a mobile robot might experience while navigating in the nursing home environment of Figure 1(b). To visualize the distribution we sample a set of poses (also called particles) according to the distribution and plot the particles on the map. The distribution is unimodal and the probability mass is mostly concentrated in a small area. Figure 4(b) shows a very different kind of belief: probability mass is spread over a wide area, there are multiple modes, and the locations of particles bear little relationship to the map. It would be difficult to find a sequence of actions and observations that would result in such a belief.

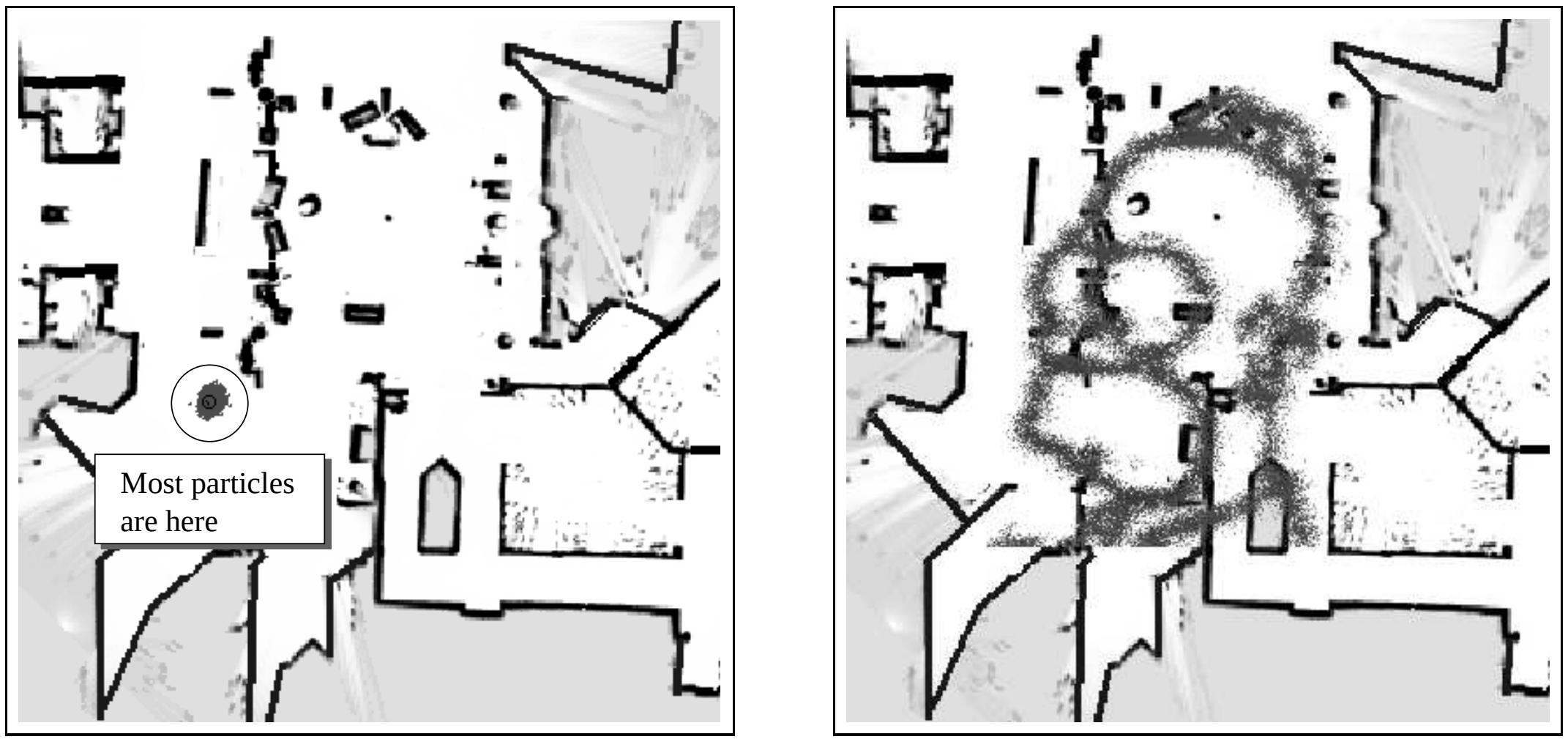

(a) A common belief

(b) An unlikely belief

**Figure 4**: Two example probability distributions over robot pose. The small black dots are particles drawn from the distribution over discrete grid positions. On the left is a distribution where the robot's location is relatively certain; this kind of compact, unimodal distribution is very common in robot navigation. On the right is a very different, implausible distribution. The right hand distribution is sufficiently unlikely that we can afford to ignore it; even if we are unable to distinguish this belief from some other belief and as a result fail to identify its optimal action, the quality of our controller will be unaffected.

If real-world beliefs have few degrees of freedom, then they should be concentrated near a low-dimensional subset of the high-dimensional belief space—that is, the beliefs experienced by the controller should lie near a structured, low-dimensional surface embedded in the belief space. If we can find this surface, we will have a representation of the belief state in terms of a small set of bases or features. One benefit of such a representation is that we will need to plan only in terms of the small set of features: finding value functions in low-dimensional spaces is typically easier than finding value functions in high-dimensional spaces.

There are two potential disadvantages to this sort of representation. The first is that it contains an approximation: we are no longer finding the complete, optimal POMDP policy. Instead (as suggested in Figure 5) we are trying to find representations of the belief which are rich enough to allow good control but which are also sufficiently parsimonious to make the planning problem tractable. The second disadvantage is a technical one: because we are making a nonlinear transformation of the belief space, POMDP planning algorithms which assume a convex value function will no longer work. We discuss this problem in more detail in Section 6.

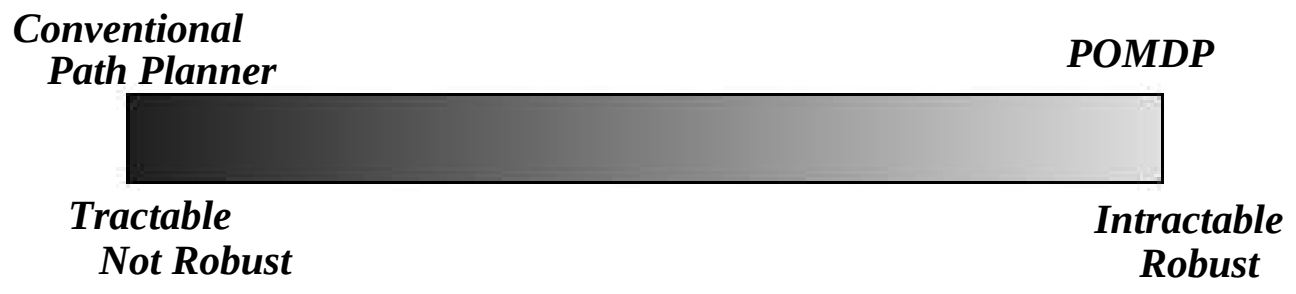

**Figure 5**: The most useful planner lies somewhere on a continuum between the MDP-style approximations and the full POMDP solution.

## 3. Dimensionality Reduction

In order to find a low-dimensional representation of our beliefs, we will use statistical dimensionality reduction algorithms (Cox & Cox, 1994). These algorithms search for a projection from our original high-dimensional representation of our beliefs to a lower-dimensional compact representation. That is, they search for a low-dimensional surface, embedded in the high-dimensional belief space, which passes near all of the sample beliefs. If we consider the evolution of beliefs from a POMDP as a trajectory inside the belief space, then our assumption is that trajectories for most large, real world POMDPs lie near a low-dimensional surface embedded in the belief space. Figure 6 depicts an example low-dimensional surface embedded in the belief space of the three-state POMDP described in the previous section.

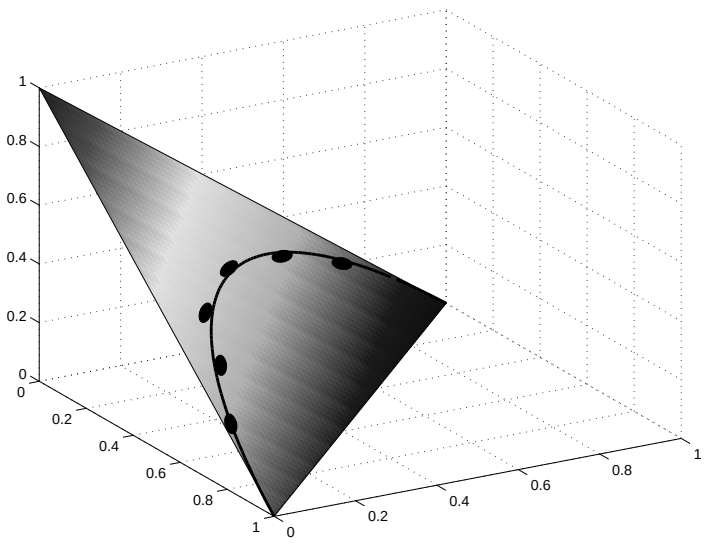

**Figure 6**: A one-dimensional surface (black line) embedded in a two-dimensional belief space (gray triangle). Each black dot represents a single belief probability distribution experienced by the controller. The beliefs all lie near the low-dimensional surface.

Ideally, dimensionality reduction involves no information loss—all aspects of the data can be recovered equally well from the low-dimensional representation as from the high-dimensional one. In practice, though, we will see that we can use lossy representations of the belief (that is, representations that may not allow the original data or beliefs to be recovered without error) and still get good control. But, we will also see that finding such representations of probability distributions will require a careful trade-off between preserving important aspects of the distributions and using as few dimensions as possible. We can measure the quality of our representation by penalizing reconstruction errors with a loss function (Collins et al., 2002). The loss function provides a quantitative way to measure errors in representing the data, and different loss functions will result in different low-dimensional representations.

### Principal Components Analysis

One of the most common forms of dimensionality reduction is Principal Components Analysis (Joliffe, 1986). Given a set of data, PCA finds the linear lower-dimensional representation of the data such that the variance of the reconstructed data is preserved. Intuitively, PCA finds a low-dimensional hyperplane such that, when we project our data onto the hyperplane, the variance of our data is changed as little as possible. A transformation that preserves variance seems appealing because it will maximally preserve our ability to distinguish between beliefs that are far apart in Euclidean norm. As we will see below, however, Euclidean norm is not the most appropriate way to measure distance between beliefs when our goal is to preserve the ability to choose good actions.

We first assume we have a data set of $n$ beliefs $\{b_1, \ldots, b_n\} \in \mathbb{B}$, where each belief $b_i$ is in $\mathbb{B}$, the high-dimensional belief space. We write these beliefs as column vectors in a matrix $B = [b_1| \ldots |b_n]$, where $B \in \mathbb{R}^{|\mathcal{S}| \times n}$. We use PCA to compute a low-dimensional representation of the beliefs by factoring $B$ into the matrices $U$ and $\tilde{B}$,

$$B = U\tilde{B}^T. \tag{1}$$

In equation (1), $U \in \mathbb{R}^{|\mathcal{S}| \times l}$ corresponds to a matrix of bases that span the low-dimensional space of $l < |\mathcal{S}|$ dimensions. $\tilde{B} \in \mathbb{R}^{n \times l}$ represents the data in the low-dimensional space.[2] From a geometric perspective, $U$ comprises a set of bases that span a hyperplane $\tilde{\mathbb{B}}$ in the high-dimensional space of $\mathbb{B}$; $\tilde{B}$ are the co-ordinates of the data on that hyperplane. If no hyperplane of dimensionality $l$ exists that contains the data exactly, PCA will find the surface of the given dimensionality that best preserves the variance of the data, after projecting the data onto that hyperplane and then reconstructing it. Minimizing the change in variance between the original data $B$ and its reconstruction $U\tilde{B}^T$ is equivalent to minimizing the sum of squared error loss:

$$L(B, U, \tilde{B}) = \|B - U\tilde{B}^T\|_F^2. \tag{2}$$

### PCA Performance

Figure 7 shows a toy problem that we can use to evaluate the success of PCA at finding low-dimensional representations. The abstract model has a two-dimensional state space: one dimension of position along one of two circular corridors, and one binary variable that determines which corridor we are in. States $s_1 \ldots s_{100}$ inclusive correspond to one corridor, and states $s_{101} \ldots s_{200}$ correspond to the other. The reward is at a known position that is different in each corridor; therefore, the agent needs to discover its corridor, move to the

2. Many descriptions of PCA are based on a factorization $USV^T$, with $U$ and $V$ column-orthonormal and $S$ diagonal. We could enforce a similar constraint by identifying $\tilde{B} = VS$; in this case the columns of $U$ would have to be orthonormal while those of $\tilde{B}$ would have to be orthogonal.

appropriate position, and declare it has arrived at the goal. When the goal is declared the system resets (regardless of whether the agent is actually at the goal). The agent has 4 actions: `left`, `right`, `sense_corridor`, and `declare_goal`. The observation and transition probabilities are given by discretized von Mises distributions (Mardia & Jupp, 2000; Shatkay & Kaelbling, 2002), an exponential family distribution defined over $[-\pi : \pi)$. The von Mises distribution is a "wrapped" analog of a Gaussian; it accounts for the fact that the two ends of the corridor are connected. Because the sum of two von Mises variates is another von Mises variate, and because the product of two von Mises likelihoods is a scaled von Mises likelihood, we can guarantee that the true belief distribution is always a von Mises distribution over each corridor after each action and observation.

This instance of the problem consists of 200 states, with 4 actions and 102 observations. Actions 1 and 2 move the controller left and right (with some von Mises noise) and action 3 returns an observation that uniquely and correctly identifies which half of the maze the agent is in (the top half or the bottom half). Observations returned after actions 1 and 2 identify the current state modulo 100: the probability of each observation is a von Mises distribution with mean equal to the true state (modulo 100). That is, these observations indicate approximately where the agent is horizontally.

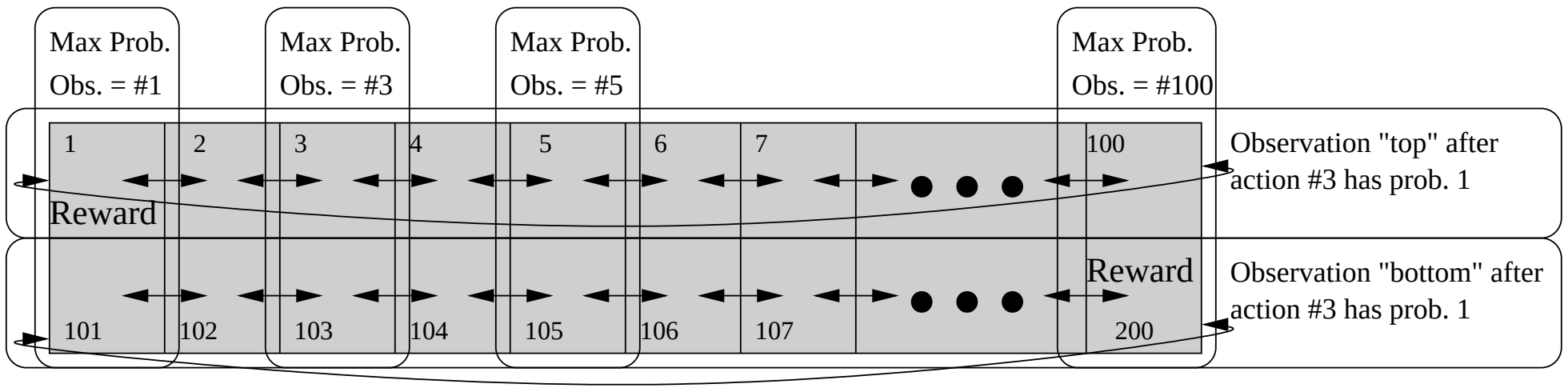

**Figure 7**: The toy maze of 200 states.

This maze is interesting because it is relatively large by POMDP standards (200 states) and contains a particular kind of uncertainty—the agent *must* use action 3 at some point to uniquely identify which half of the maze it is in; the remaining actions result in observations that contain no information about which corridor the agent is in. This problem is too large to be solved by conventional POMDP value iteration, but structured such that heuristic policies will also perform poorly.

We collected a data set of 500 beliefs and assessed the performance of PCA on beliefs from this problem. The data were collected using a hand-coded controller, alternating at random between exploration actions and the MDP solution, taking as the current state the maximum-likelihood state of the belief. Figure 8 shows 4 sample beliefs from this data set. Notice that each of these beliefs is essentially two discretized von Mises distributions with different weights, one for each half of the maze. The starting belief state is the left-most distribution in Figure 8: equal probability on the top and bottom corridors, and position along the corridor following a discretized von Mises distribution with concentration parameter 1.0 (meaning that $p(\text{state})$ falls to $1/e$ of its maximum value when we move $1/4$ of the way around the corridor from the most likely state).

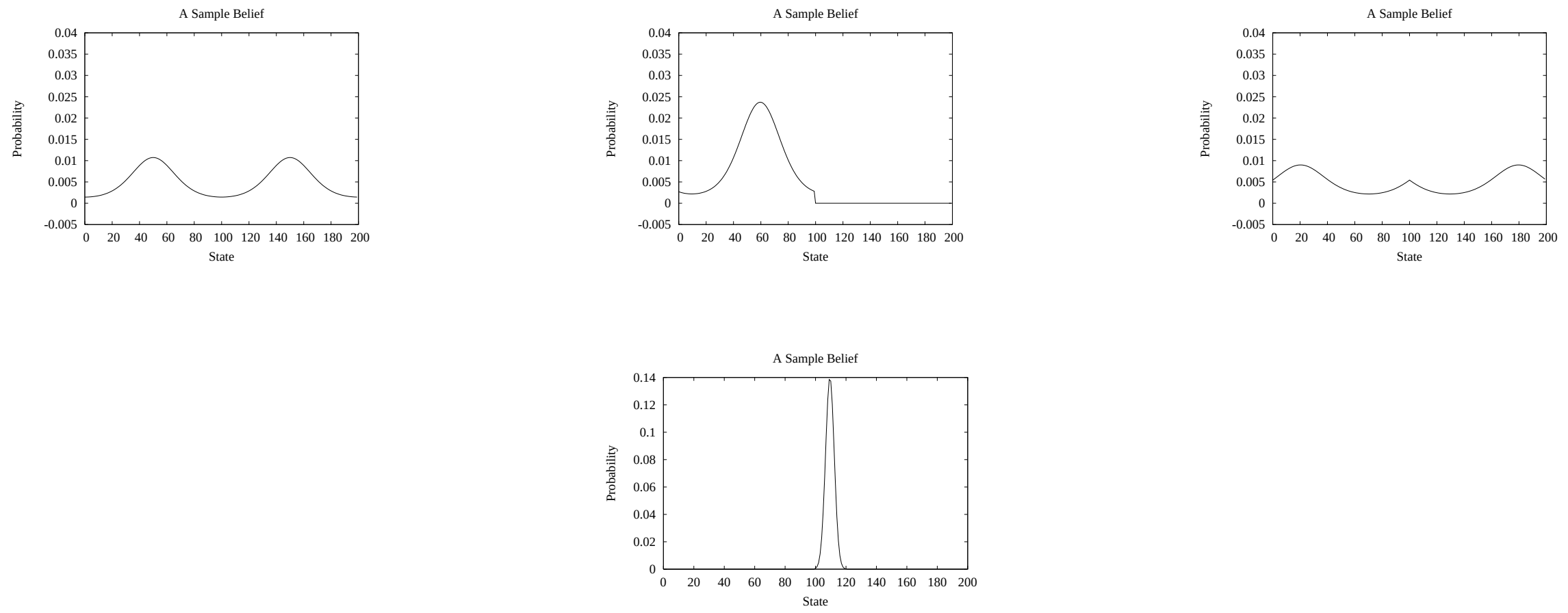

**Figure 8**: Sample beliefs for the toy problem, from a sample set of 500, at different (non-contiguous) points in time. The left-most belief is the initial belief state.

Figure 9 examines the performance of PCA on representing the beliefs in this data set by computing the average error between the original beliefs $B$ and their reconstructions $U\tilde{B}$.[3] In Figure 9(a) we see the average squared error (squared $L_2$) compared to the number of bases, and in Figure 9(b), we see the average Kullbach-Leibler (KL) divergence. The KL divergence between a belief $b$ and its reconstruction $r = U\tilde{b}$ from the low-dimensional representation $\tilde{b}$ is given by

$$KL(b \,\|\, r) = \sum_{i=1}^{|\mathcal{S}|} b(s_i) \ln \left( \frac{b(s_i)}{r(s_i)} \right) \tag{3}$$

Minimizing the squared $L_2$ error is the explicit objective of PCA, but the KL divergence is a more appropriate measure of how much two probability distributions differ. [4]

Unfortunately, PCA performs poorly at representing probability distributions. Despite the fact that probability distributions in the collected data set have only 3 degrees of freedom, the reconstruction error remains relatively high until somewhere between 10 and 15 basis functions. If we examine the reconstruction of a sample belief, we can see the kinds of errors that PCA is making. Figure 10 shows a sample belief (the solid line) and its reconstruction (the dotted line). Notice that the reconstructed belief has some strange artifacts: it contains ringing (multiple small modes), and also is negative in some regions. PCA is a purely geometric process; it has no notion of the original data as probability distributions, and is therefore free to generate reconstructions of the data that contain negative numbers or do not sum to 1.

3. We use a popular implementation of PCA based on the Golub-Reinsche algorithm (Golub & Reinsch, 1970) available through the GNU Scientific Library (Galassi, Davies, Theiler, Gough, Jungman, Booth, & Rossi, 2002).
4. Note that before computing the KL divergence between the reconstruction and the original belief, we shift the reconstruction to be non-negative, and rescale it to sum to 1.

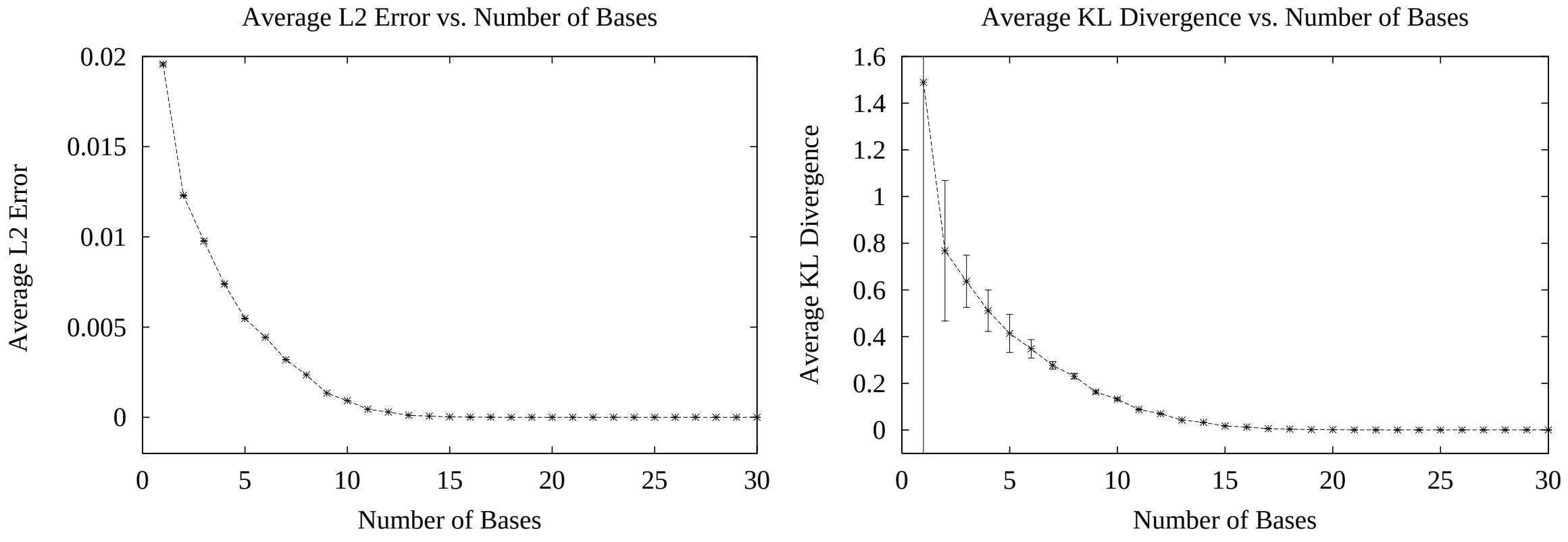

(a) Average Squared L-2 Error

(b) Average KL Divergence

**Figure 9**: The average error between the original sample set $B$ and the reconstructions $U\tilde{B}$. (a) Squared $L_2$ error, explicitly minimized by PCA, and (b) the KL divergence. The error bars represent the standard deviation from the mean of the error over the 500 beliefs.

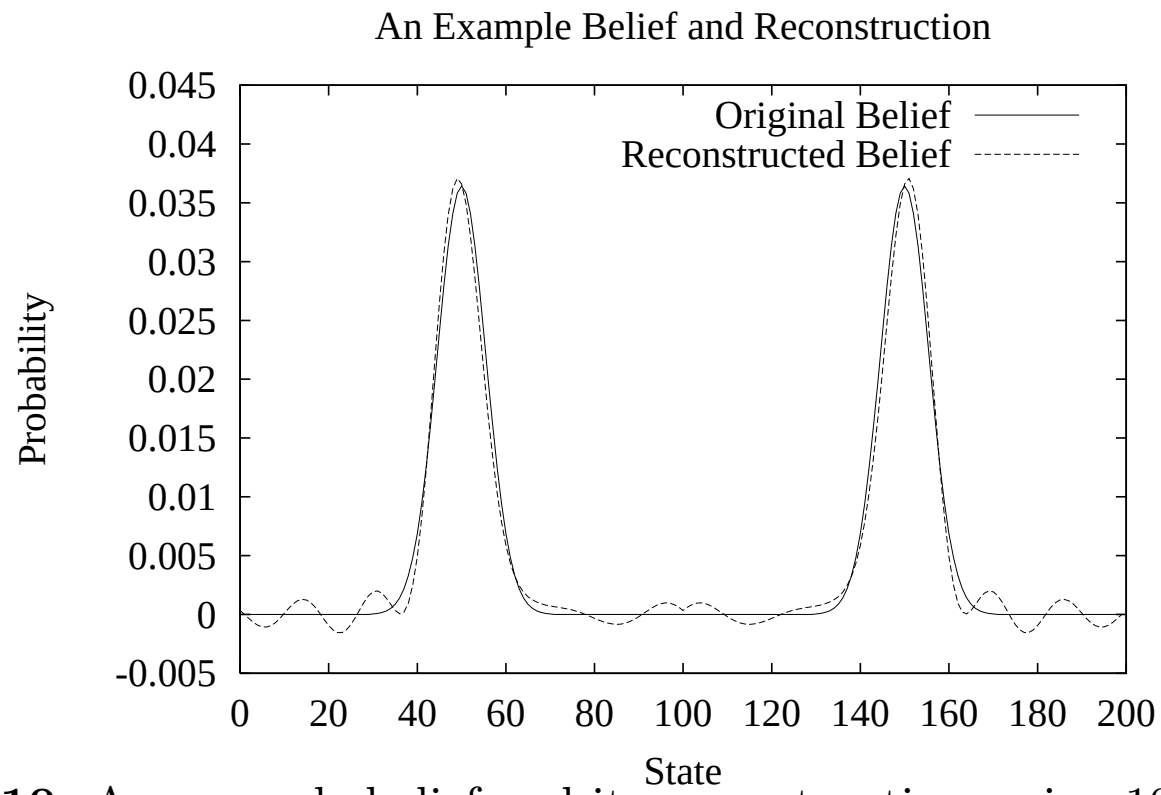

**Figure 10**: An example belief and its reconstruction, using 10 bases.

Notice also that the PCA process is making its most significant errors in the low-probability regions of the belief. This is particularly unfortunate because real-world probability distributions tend to be characterized by compact masses of probability, surrounded by large regions of zero probability (e.g., Figure 4a). What we therefore need to do is modify PCA to ensure the reconstructions are probability distributions, and improve the representation of sparse probability distributions by reducing the errors made on low-probability events.

The question to be answered is what loss functions are available instead of the sum of squared errors, as in equation (2). We would like a loss function that better reflects the need to represent probability distributions.

## 4. Exponential Family PCA

The conventional view of PCA is a geometric one, finding a low-dimensional projection that minimizes the squared-error loss. An alternate view is a probabilistic one: if the data consist of samples drawn from a probability distribution, then PCA is an algorithm for finding the parameters of the generative distribution that maximize the likelihood of the data. The squared-error loss function corresponds to an assumption that the data is generated from a Gaussian distribution. Collins et al. (2002) demonstrated that PCA can be generalized to a range of loss functions by modeling the data with different exponential families of probability distributions such as Gaussian, binomial, or Poisson. Each such exponential family distribution corresponds to a different loss function for a variant of PCA, and Collins et al. (2002) refer to the generalization of PCA to arbitrary exponential family data-likelihood models as "Exponential family PCA" or E-PCA.

An E-PCA model represents the reconstructed data using a low-dimensional weight vector $\tilde{b}$, a basis matrix $U$, and a *link function* $f$:

$$b \approx f(U\tilde{b}) \tag{4}$$

Each E-PCA model uses a different link function, which can be derived from its data likelihood model (and its corresponding error distribution and loss function). The link function is a mapping from the data space to another space in which the data can be linearly represented.

The link function $f$ is the mechanism through which E-PCA generalizes dimensionality reduction to non-linear models. For example, the identity link function corresponds to Gaussian errors and reduces E-PCA to regular PCA, while the sigmoid link function corresponds to Bernoulli errors and produces a kind of "logistic PCA" for 0-1 valued data. Other nonlinear link functions correspond to other non-Gaussian exponential families of distributions.

We can find the parameters of an E-PCA model by maximizing the log-likelihood of the data under the model, which has been shown (Collins et al., 2002) to be equivalent to minimizing a generalized Bregman divergence

$$B_{F^*}(b \,\|\, U\tilde{b}) \quad = \quad F(U\tilde{b}) - b \cdot U\tilde{b} + F^*(b) \tag{5}$$

between the low-dimensional and high-dimensional representations, which we can solve using convex optimization techniques. (Here $F$ is a convex function whose derivative is $f$, while $F^*$ is the convex dual of $F$. We can ignore $F^*$ for the purpose of minimizing equation 5 since the value of $b$ is fixed.) The relationship between PCA and E-PCA through link functions is reminiscent of the relationship between linear regression and Generalized Linear Models (McCullagh & Nelder, 1983).

To apply E-PCA to belief compression, we need to choose a link function which accurately reflects the fact that our beliefs are probability distributions. If we choose the link function

$$f(U\tilde{b}) = e^{U\tilde{b}} \tag{6}$$

then it is not hard to verify that $F(U\tilde{b}) = \sum e^{U\tilde{b}}$ and $F^*(b) = b \cdot \ln b - \sum b$. So, equation 5 becomes

$$B_{F^*}(b \,\|\, U\tilde{b}) = \sum e^{U\tilde{b}} - b \cdot U\tilde{b} + b \cdot \ln b - \sum b \tag{7}$$

If we write $\hat{b} = f(U\tilde{b})$, then equation 7 becomes

$$B_{F^*}(b \,\|\, U\tilde{b}) = b \cdot \ln b - b \cdot \ln \hat{b} + \sum \hat{b} - \sum b = UKL(b \,\|\, \hat{b})$$

where $UKL$ is the unnormalized KL divergence. Thus, choosing the exponential link function (6) corresponds to minimizing the unnormalized KL divergence between the original belief and its reconstruction. This loss function is an intuitively reasonable choice for measuring the error in reconstructing a probability distribution.[5] The exponential link function corresponds to a Poisson error model for each component of the reconstructed belief.

Our choice of loss and link functions has two advantages: first, the exponential link function constrains our low-dimensional representation $e^{U\tilde{b}}$ to be positive. Second, our error model predicts that the variance of each belief component is proportional to its expected value. Since PCA makes significant errors close to 0, we wish to increase the penalty for errors in small probabilities, and this error model accomplishes that.

If we compute the loss for all $b_i$, ignoring terms that depend only on the data $b$, then[6]

$$L(B, U, \tilde{B}) = \sum_{i=1}^{|B|} \left( e^{U\tilde{b}_i} - b_i \cdot U\tilde{b}_i \right). \tag{8}$$

The introduction of the link function raises a question: instead of using the complex machinery of E-PCA, could we just choose some non-linear function to project the data into a space where it is linear, and then use conventional PCA? The difficulty with this approach is of course identifying that function; in general, good link functions for E-PCA are not related to good nonlinear functions for application before regular PCA. So, while it might appear reasonable to use PCA to find a low-dimensional representation of the *log* beliefs, rather than use E-PCA with an *exponential* link function to find a representation of the beliefs directly, this approach performs poorly because the surface is only locally well-approximated by a log projection. E-PCA can be viewed as minimizing a weighted least-squares that chooses the distance metric to be appropriately local. Using conventional PCA over log beliefs also performs poorly in situations where the beliefs contain extremely small or zero probability entries.

5. If we had chosen the link function $e^{U\tilde{b}} / \sum e^{U\tilde{b}}$ we would have arrived at the normalized KL divergence, which is perhaps an even more intuitively reasonable way to measure the error in reconstructing a probability distribution. This more-complicated link function would have made it more difficult to derive the Newton equations in the following pages, but not impossible; we have experimented with the resulting algorithm and found that it produces qualitatively similar results to the algorithm described here. Using the normalized KL divergence does have one advantage: it can allow us to get away with one fewer basis function during planning, since for unnormalized KL divergence the E-PCA optimization must learn a basis which can explicitly represent the normalization constant.
6. E-PCA is related to Lee and Seung's (1999) non-negative matrix factorization. One of the NMF loss functions presented by Lee and Seung (1999) penalizes the KL-divergence between a matrix and its reconstruction, as we do in equation 8; but, the NMF loss does not incorporate a link function and so is not an E-PCA loss. Another NMF loss function presented by Lee and Seung (1999) penalizes squared error but constrains the factors to be nonnegative; the resulting model is an example of a $(GL)^2M$, a generalization of E-PCA described by Gordon (2003).

### Finding the E-PCA Parameters

Algorithms for conventional PCA are guaranteed to converge to a unique answer independent of initialization. In general, E-PCA does not have this property: the loss function (8) may have multiple distinct local minima. However, the problem of finding the best $\tilde{B}$ given $B$ and $U$ is convex; convex optimization problems are well studied and have unique global solutions (Rockafellar, 1970). Similarly, the problem of finding the best $U$ given $B$ and $\tilde{B}$ is convex. So, the possible local minima in the joint space of $U$ and $\tilde{B}$ are highly constrained, and finding $U$ and $\tilde{B}$ does not require solving a general non-convex optimization problem.

Gordon (2003) describes a fast, Newton's Method approach for computing $U$ and $\tilde{B}$ which we summarize here. This algorithm is related to Iteratively Reweighted Least Squares, a popular algorithm for generalized linear regression (McCullagh & Nelder, 1983). In order to use Newton's Method to minimize equation (8), we need its derivative with respect to $U$ and $\tilde{B}$:

$$\begin{aligned} \frac{\partial}{\partial U} L(B, U, \tilde{B}) &= \frac{\partial}{\partial U} e^{(U\tilde{B})} - \frac{\partial}{\partial U} B \circ U\tilde{B} && (9) \\ &= e^{(U\tilde{B})}\tilde{B}^T - B\tilde{B}^T && (10) \\ &= (e^{(U\tilde{B})} - B)\tilde{B}^T && (11) \end{aligned}$$

and

$$\begin{aligned} \frac{\partial}{\partial \tilde{B}} L(B, U, \tilde{B}) &= \frac{\partial}{\partial \tilde{B}} e^{(U\tilde{B})} - \frac{\partial}{\partial \tilde{B}} B \circ U\tilde{B} && (12) \\ &= U^T e^{(U\tilde{B})} - U^T B && (13) \\ &= U^T (e^{(U\tilde{B})} - B). && (14) \end{aligned}$$

If we set the right hand side of equation (14) to zero, we can iteratively compute $\tilde{B}_{\cdot j}$, the $j^{th}$ column of $\tilde{B}$, by Newton's method. Let us set $q(\tilde{B}_{\cdot j}) = U^T(e^{(U\tilde{B}_{\cdot j})} - B_{\cdot j})$, and linearize about $\tilde{B}_{\cdot j}$ to find roots of $q()$. This gives

$$\begin{aligned} \tilde{B}^{new}_{\cdot j} &= \tilde{B}_{\cdot j} - \frac{q(\tilde{B}_{\cdot j})}{q'(\tilde{B}_{\cdot j})} && (15) \\ \tilde{B}^{new}_{\cdot j} - \tilde{B}_{\cdot j} &= -\frac{U^T(e^{(U\tilde{B}_{\cdot j})} - B_{\cdot j})}{q'(\tilde{B}_{\cdot j})} && (16) \end{aligned}$$

Note that equation 15 is a formulation of Newton's method for finding roots of $q$, typically written as

$$x_{n+1} = x_n - \frac{f(x_n)}{f'(x_n)}. \quad (17)$$

We need an expression for $q'$:

$$\begin{aligned} \frac{\partial q}{\partial \tilde{B}_{\cdot j}} &= \frac{\partial}{\partial \tilde{B}_{\cdot j}} U^T(e^{(U\tilde{B}_{\cdot j})} - B_{\cdot j}) && (18) \\ &= \frac{\partial}{\partial \tilde{B}_{\cdot j}} U^T e^{(U\tilde{B}_{\cdot j})} && (19) \\ &= U^T D_j U && (20) \end{aligned}$$

We define $D_j$ in terms of the `diag` operator that returns a diagonal matrix:

$$D_j = \texttt{diag}(e^{U\tilde{B}_{\cdot j}}), \tag{21}$$

where

$$\texttt{diag}(b) = \begin{bmatrix} b(s_0) & \dots & 0 \\ \vdots & \ddots & \vdots \\ 0 & \dots & b(s_{|\mathcal{S}|}) \end{bmatrix}. \tag{22}$$

Combining equation (15) and equation (20), we get

$$(U^T D_j U)(\tilde{B}^{new}_{\cdot j} - \tilde{B}_{\cdot j}) = U^T(B_{\cdot j} - e^{U\tilde{B}_{\cdot j}}) \tag{23}$$

$$U^T D_j U \tilde{B}^{new}_{\cdot j} = (U^T D_j U)\tilde{B}_{\cdot j} + U^T D_j D_j^{-1}(B_{\cdot j} - e^{U\tilde{B}_{\cdot j}}) \tag{24}$$

$$= U^T D_j (U\tilde{B}_{\cdot j} + D_j^{-1}(B_{\cdot j} - e^{U\tilde{B}_{\cdot j}}), \tag{25}$$

which is a weighted least-squares problem that can be solved with standard linear algebra techniques. In order to ensure that the solution is numerically well-conditioned, we typically add a regularizer to the divisor, as in

$$\tilde{B}^{new}_{\cdot j} = \frac{U^T D_j (U\tilde{B}_{\cdot j} + D_j^{-1}(B_{\cdot j} - e^{U\tilde{B}_{\cdot j}})}{(U^T D_j U + 10^{-5} I_l)}. \tag{26}$$

where $I_l$ is the $l \times l$ identity matrix. Similarly, we can compute a new $U$ by computing $U_{i\cdot}$, the $i^{th}$ row of $U$, as

$$U^{new}_{i\cdot} = \frac{(U_{i\cdot}\tilde{B} + (B_{i\cdot} - e^{U_{i\cdot}\tilde{B}})D_i^{-1})D_i\tilde{B}^T}{(\tilde{B}D_i\tilde{B}^T + 10^{-5} I_l)}. \tag{27}$$

### The E-PCA Algorithm

We now have an algorithm for automatically finding a good low-dimensional representation $\tilde{\mathbb{B}}$ for the high-dimensional belief set $\mathbb{B}$. This algorithm is given in Table 1; the optimization is iterated until some termination condition is reached, such as a finite number of iterations, or when some minimum error $\epsilon$ is achieved.

The steps 7 and 9 raise one issue. Although solving for each row of $U$ or column of $\tilde{B}$ separately is a convex optimization problem, solving for the two matrices simultaneously is not. We are therefore subject to potential local minima; in our experiments we did not find this to be a problem, but we expect that we will need to find ways to address the local minimum problem in order to scale to even more complicated domains.

Once the bases $U$ are found, finding the low-dimensional representation of a high-dimensional belief is a convex problem; we can compute the best answer by iterating equation (26). Recovering a full-dimensional belief $b$ from the low-dimensional representation $\tilde{b}$ is also very straightforward:

$$x = e^{U\tilde{b}}. \tag{28}$$

Our definition of PCA does not explicitly factor the data into $U$, $S$ and $\tilde{B}$ as many presentations do. In this three-part representation of PCA, $S$ contains the singular values

Collect a set of sample beliefs from the high-dimensional belief space<br>
Assemble the samples into the data matrix $B = [b_1| \dots |b_{|B|}]$<br>
Choose an appropriate loss function, $L(B, U, \tilde{B})$<br>
Fix an initial estimate for $\tilde{B}$ and $U$ randomly<br>
do<br>
    For each column $\tilde{B}_{\cdot j} \in \tilde{B}$,<br>
        Compute $\tilde{B}^{new}_{\cdot j}$ using current $U$ estimate from equation (26)<br>
    For each row $U_{i\cdot} \in U$,<br>
        Compute $U^{new}_{i\cdot}$ using new $\tilde{B}$ estimate from equation (27)<br>
while $L(B, U, \tilde{B}) > \epsilon$

**Table 1**: The E-PCA Algorithm for finding a low-dimensional representation of a POMDP, including Gordon's Newton's method (2003).

of the decomposition, and $U$ and $\tilde{B}$ are orthonormal. We use the two-part representation $B \approx f(U\tilde{B})$ because there is no quantity in the E-PCA decomposition which corresponds to the singular values in PCA. As a result, $U$ and $\tilde{B}$ will not in general be orthonormal. If desired, though, it is possible to orthonormalize $U$ as an additional step after optimization using conventional PCA and adjust $\tilde{B}$ accordingly.

## 5. E-PCA Performance

Using the loss function from equation (8) with the iterative optimization procedure described by equation (26) and equation (27) to find the low-dimensional factorization, we can look at how well this dimensionality-reduction procedure performs on some POMDP examples.

### Toy Problem

Recall from Figure 9 that we were unable to find good representations of the data with fewer than 10 or 15 bases, even though our domain knowledge indicated that the data had 3 degrees of freedom (horizontal position of the mode along the corridor, concentration about the mode, and probability of being in the top or bottom corridor). Examining one of the sample beliefs in Figure 10, we saw that the representation was worst in the low-probability regions. We can now take the same data set from the toy example, use E-PCA to find a low-dimensional representation and compare the performance of PCA and E-PCA. Figure 11(a) shows that E-PCA is substantially more efficient at representing the data, as we see the KL divergence falling very close to 0 after 4 bases. Additionally, the squared $L_2$ error at 4 bases is $4.64 \times 10^{-4}$. (We need 4 bases for perfect reconstruction, rather than 3, since we must include a constant basis function. The small amount of reconstruction error with 4 bases remains because we stopped the optimization procedure before it fully converged.)

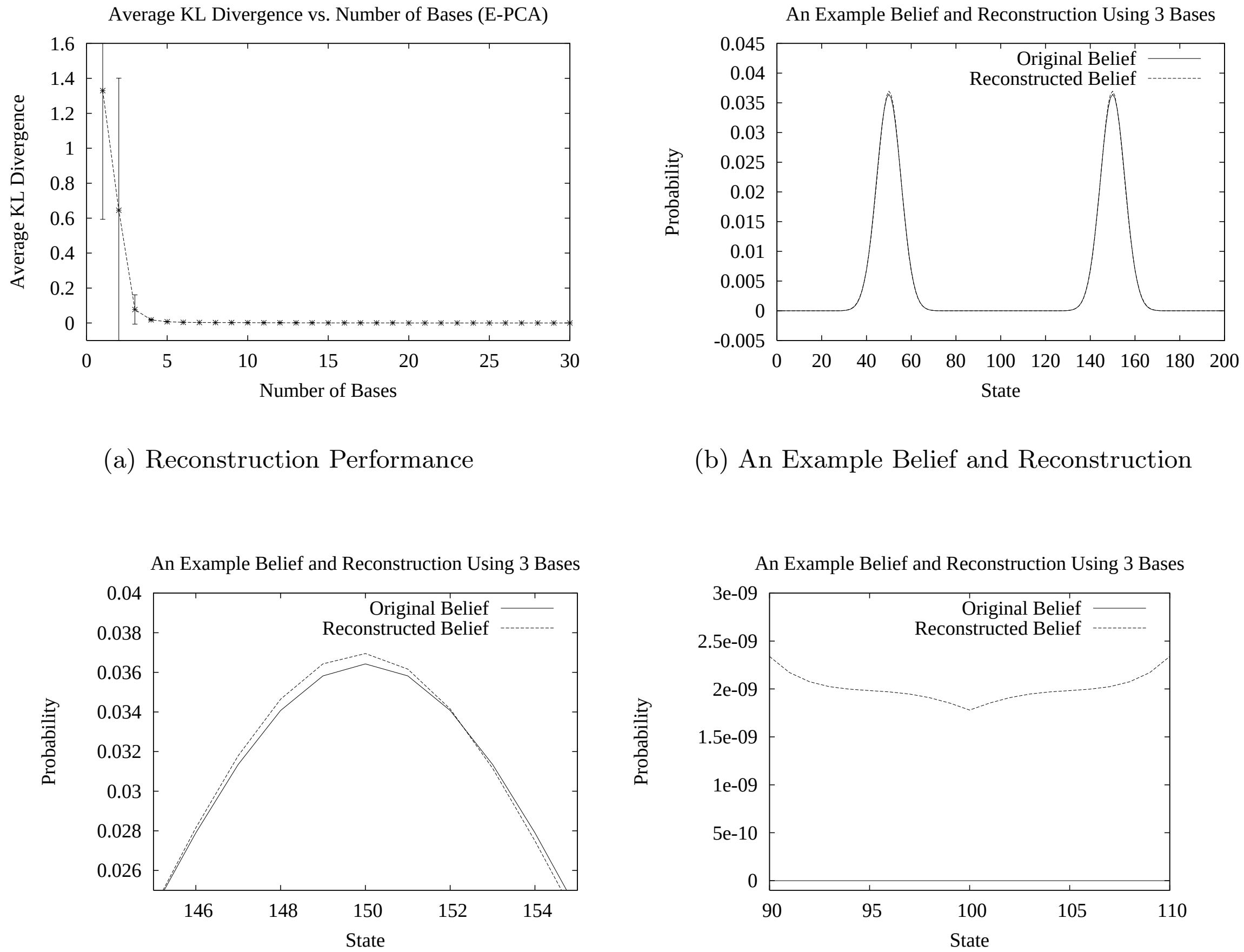

(a) Reconstruction Performance

(b) An Example Belief and Reconstruction

(c) The Belief and Reconstruction Near the Peak

(d) The Belief and Reconstruction In Low-Probability Region

**Figure 11**: (a) The average KL divergence between the original sample set and the reconstructions. The KL divergence is 0.018 after 4 bases. The error bars represent the standard deviation from the mean over the 500 beliefs. (b) The same example belief from Figure 10 and its reconstruction using 3 bases. The reconstruction shows small errors at the peak of each mode. Not shown is the reconstruction using 4 bases, in which the original belief and its reconstruction are indistinguishable to the naked eye. (c) and (d) show fine detail of the original belief and the reconstruction in two parts of the state space. Although the reconstruction is not perfect, in the low-probability area, we see that the error is approximately $2 \times 10^{-9}$.

Figure 11(b) shows the E-PCA reconstruction of the same example belief as in Figure 10. We see that many of the artifacts present in the PCA reconstruction are absent. Using only 3 bases, we see that the E-PCA reconstruction is already substantially better than PCA using 10 bases, although there are some small errors at the peaks (e.g., Figure 11c) of the two modes. (Using 4 bases, the E-PCA reconstruction is indistinguishable to the naked eye from the original belief.) This kind of accuracy for both 3 and 4 bases is typical for this data set.

### Robot Beliefs

Although the performance of E-PCA on finding good representations of the abstract problem is compelling, we would ideally like to be able to use this algorithm on real-world problems, such as the robot navigation problem in Figure 2. Figures 12 and 13 show results from two such robot navigation problems, performed using a physically-realistic simulation (although with artificially limited sensing and dead-reckoning). We collected a sample set of 500 beliefs by moving the robot around the environment using a heuristic controller, and computed the low-dimensional belief space $\tilde{B}$ according to the algorithm in Table 1. The full state space is $47.7m \times 17m$, discretized to a resolution of $1m \times 1m$ per pixel, for a total of 799 states. Figure 12(a) shows a sample belief, and Figure 12(b) the reconstruction using 5 bases. In Figure 12(c) we see the average reconstruction performance of the E-PCA approach, measured as average KL-divergence between the sample belief and its reconstruction. For comparison, the performance of both PCA and E-PCA are plotted. The E-PCA error falls to 0.02 at 5 bases, suggesting that 5 bases are sufficient for good reconstruction. This is a very substantial reduction, allowing us to represent the beliefs in this problem using only 5 parameters, rather than 799 parameters. Notice that many of the states lie in regions that are "outside" the map; that is, states that can never receive probability mass were not removed. While removing these states would be a trivial operation, the E-PCA is correctly able to do so automatically.

In Figure 13, similar results are shown for a different environment. A sample set of 500 beliefs was again collected using a heuristic controller, and the low-dimensional belief space $\tilde{B}$ was computed using the E-PCA. The full state space is $53.6m \times 37.9m$, with a resolution of $.5m \times .5m$ per pixel. An example belief is shown in Figure 13(a), and its reconstruction using 6 bases is shown Figure 13(b). The reconstruction performance as measured by the average KL divergence is shown in Figure 13(c); the error falls very close to 0 around 6 bases, with minimal improvement thereafter.

## 6. Computing POMDP policies

The Exponential-family Principal Components Analysis model gives us a way to find a low-dimensional representation of the beliefs that occur in any particular problem. For the two real-world navigation problems we have tried, the algorithm proved to be effective at finding very low-dimensional representations, showing reductions from $\approx 800$ states and $\approx 2,000$ states down to 5 or 6 bases. A 5 or 6 dimensional belief space will allow much more tractable computation of the value function, and so we will be able to solve much larger POMDPs than we could have solved previously.

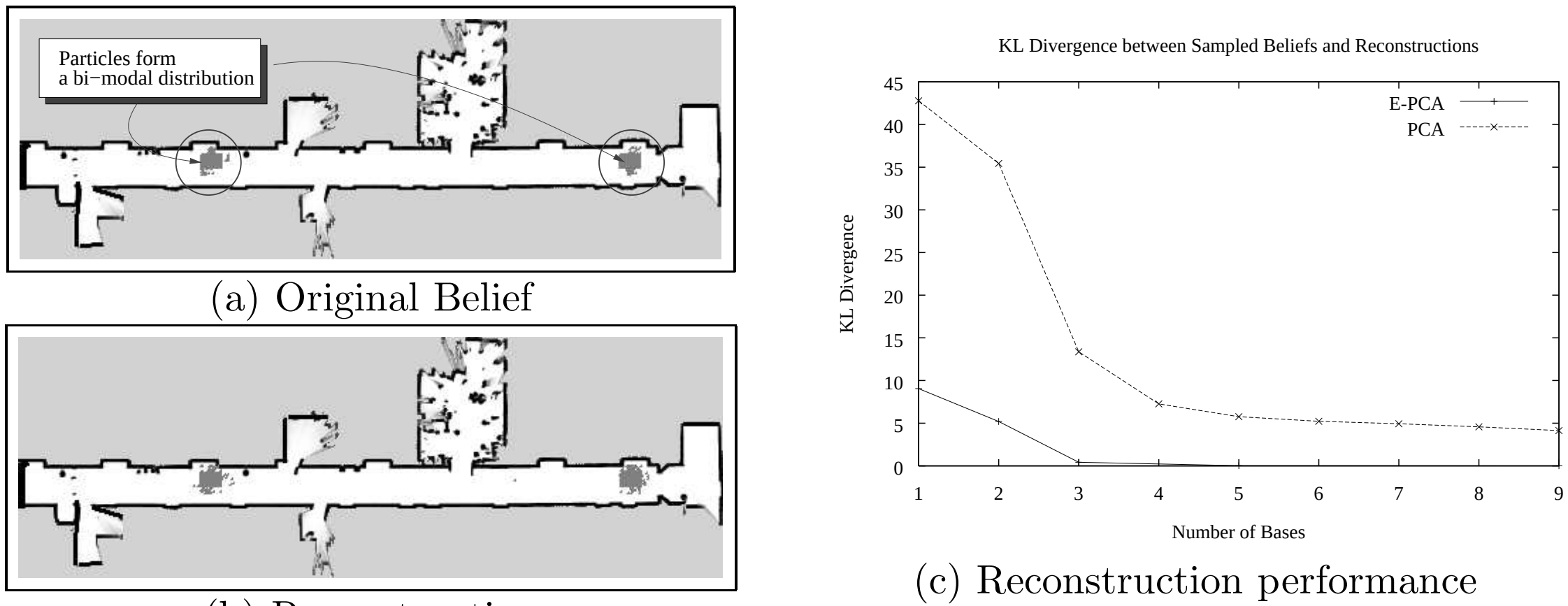

(a) Original Belief

(b) Reconstruction

(c) Reconstruction performance

**Figure 12**: (a) A sample belief for the robot navigation task. (b) The reconstruction of this belief from the learned E-PCA representation using 5 bases. (c) The average KL divergence between the sample beliefs and their reconstructions against the number of bases used. Notice that the E-PCA error falls close to 0 for 5 bases, whereas conventional PCA has much worse reconstruction error even for 9 bases, and is not improving rapidly.

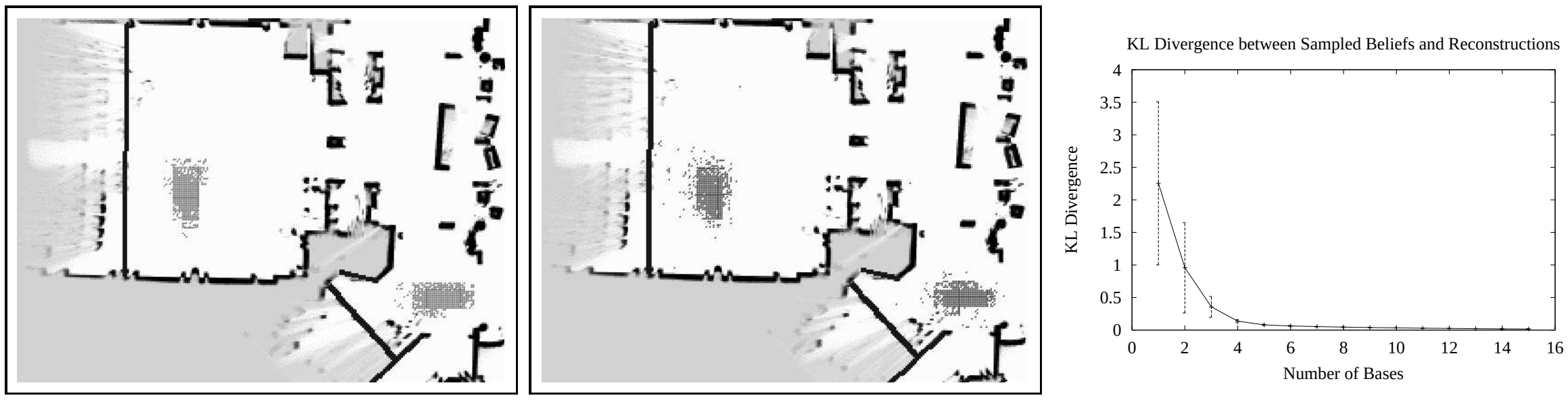

(a) A sample belief

(b) The reconstruction

(c) Average reconstruction performance

**Figure 13**: (a) A sample belief for the navigation problem in Longwood, cf. Figure 2. (b) The reconstruction from the learned E-PCA representation using 6 bases. (c) The average KL divergence between the sample beliefs and their reconstructions against the number of bases used.

Unfortunately, we can no longer use conventional POMDP value iteration to find the optimal policy given the low-dimensional set of belief space features. POMDP value iteration depends on the fact that the value function is convex over the belief space. When we compute a non-linear transformation of our beliefs to recover their coordinates on the low-dimensional belief surface, we lose the convexity of the value function (compare Figure 3 and Figure 6 to see why). As a result, the value function cannot be expressed as the supremum of a set of hyperplanes over the low-dimensional belief space.

So, instead of using POMDP value iteration, we will build a low-dimensional discrete belief space MDP and use MDP value iteration. Since we do not know the form of the value function, we will turn to function approximation. Gordon (1995) proved that the *fitted value iteration* algorithm is guaranteed to find a bounded-error approximation to a (possibly discounted) MDP's value function, so long as we use it in combination with a function approximator that is an *averager*. Averagers are function approximators which are non-expansions in max-norm; that is, they do not exaggerate errors in their training data. In our experiments below, we use regular grids as well as irregular, variable-resolution grids based on 1-nearest-neighbour discretization, represented by a set of low-dimensional beliefs $\tilde{B}^*$,

$$\tilde{B}^* = \{\tilde{b}^*_1, \tilde{b}^*_2, \ldots, \tilde{b}^*_{|\tilde{B}^*|}\}. \tag{29}$$

Both of these approximations are averagers; other averagers include linear interpolation, $k$-nearest-neighbours, and local weighted averaging. We will not focus in detail on the exact mechanism for discretizing the low-dimensional space, as this is outside the scope of this paper. The resolution of the regular grid in all cases was chosen empirically; in section 7 we describe a specific variable resolution discretization scheme that worked well empirically. The reader can consult Munos and Moore (2002) or Zhou and Hansen (2001) for more sophisticated representations.

The fitted value iteration algorithm uses the following update rule to compute a $t$-step lookahead value function $V^t$ from a $(t-1)$-step lookahead value function $V^{t-1}$:

$$V^t(\tilde{b}^*_i) = \max_a \left( \tilde{R}^*(\tilde{b}^*_i, a) + \gamma \sum_{j=1}^{|\tilde{B}^*|} \tilde{T}^*(\tilde{b}^*_i, a, \tilde{b}^*_j) \cdot V^{t-1}(\tilde{b}^*_j) \right) \tag{30}$$

Here $\tilde{R}^*$ and $\tilde{T}^*$ are approximate reward and transition functions based on the dynamics of our POMDP, the result of our E-PCA, and the finite set of low-dimensional belief samples $\tilde{B}^*$ that we are using as our function approximator. Note that in all problems described in this paper, the problem did not require discounting ($\gamma = 1$). The following sections describe how to compute the model parameters $\tilde{R}^*$ and $\tilde{T}^*$.

### Computing the Reward Function

The original reward function $R(s, a)$ represents the immediate reward of taking action $a$ at state $s$. We cannot know, given either a low-dimensional or high-dimensional belief, what the immediate reward *will* be, but we can compute the *expected* reward. We therefore represent the reward as the expected value of the immediate reward of the full model, under the current belief:

$$\tilde{R}^*(\tilde{b}^*, a) = \mathrm{E}_{b^*}(R(s, a)) \tag{31}$$

$$= \sum_{i=1}^{|\mathcal{S}|} R(s_i, a) b(s_i). \tag{32}$$

Equation (32) requires us to recover the high-dimensional belief $b$ from the low-dimensional representation $\tilde{b}^*$, as shown in equation (28).

For many problems, the reward function $\tilde{R}^*$ will have the effect of giving a low immediate reward for belief states with high entropy. That is, for many problems the planner will be driven towards beliefs that are centred on high-reward states and have low uncertainty. This property is intuitively desirable: in such beliefs the robot does not have to worry about an immediate bad outcome.

### Computing the Transition Function

Computing the low-dimensional transition function $\tilde{T}^* = p(\tilde{b}_j^*|a, \tilde{b}_i^*)$ is not as simple as computing the low-dimensional reward function $\tilde{R}^*$: we need to consider pairs of low-dimensional beliefs, $\tilde{b}_i^*$ and $\tilde{b}_j^*$. In the original high-dimensional belief space, the transition from a prior belief $b_i$ to a posterior belief $b_j$ is described by the Bayes filter equation:

$$b_j(s) \quad = \quad \alpha\, O(s, a, z) \sum_{k=1}^{|\mathcal{S}|} T(s_k, a, s) b_i(s_k) \tag{33}$$

Here $a$ is the action we selected and $z$ is the observation we saw; $T$ is the original POMDP transition probability distribution, and $O$ is the original POMDP observation probability distribution.

Equation (33) describes a deterministic transition conditioned upon a prior belief, an action and an observation. The transition to the posterior $b_j$ is stochastic when the observation is not known; that is, the transition from $b_i$ to $b_j$ occurs only when a specific $z$ is generated, and the probability of this transition is the probability of generating observation $z$. So, we can separate the full transition process into a deterministic transition to $b_a$, the belief after acting but before sensing, and a stochastic transition to $b_j$, the full posterior:

$$b_a(s) \quad = \quad \sum_{j=1}^{|\mathcal{S}|} T(s_j, a, s) b_i(s_j) \tag{34}$$

$$b_j(s) \quad = \quad \alpha\, O(s, a, z) b_a(s). \tag{35}$$

Equations 34 and 35 describe the transitions of the high-dimensional beliefs for the original POMDP. Based on these high-dimensional transitions, we can compute the transitions in our low-dimensional approximate belief space MDP. Figure 14 depicts the process. As the figure shows, we start with a low-dimensional belief $\tilde{b}_i^*$. From $\tilde{b}_i^*$ we reconstruct a high-dimensional belief $b$ according to equation (28). Then we apply an action $a$ and an observation $z$ as described in equation (34) and equation (35) to find the new belief $b'$. Once we have $b'$ we can compress it to a low-dimensional representation $\tilde{b}'$ by iterating equation (26). Finally, since $\tilde{b}'$ may not be a member of our sample $\tilde{B}^*$ of low-dimensional belief states, we map $\tilde{b}'$ to a nearby $\tilde{b}_j^* \in \tilde{B}^*$ according to our function approximator.

If our function approximator is a grid, the last step above means replacing $\tilde{b}'$ by a prototypical $\tilde{b}_j^*$ which shares its grid cell. More generally, our function approximator may represent $\tilde{b}'$ as a combination of several states, putting weight $w(\tilde{b}_j^*, \tilde{b}')$ on each $\tilde{b}_j^*$. (For example, if our approximator is $k$-nearest-neighbour, $w(\tilde{b}_j^*, \tilde{b}') = \frac{1}{k}$ for each of the closest $k$ samples in $\tilde{B}^*$.) In this case we replace the transition from $\tilde{b}_i^*$ to $\tilde{b}'$ with several transitions, each from $\tilde{b}_i^*$ to some $\tilde{b}_j^*$, and scale the probability of each one by $w(\tilde{b}_j^*, \tilde{b}')$.

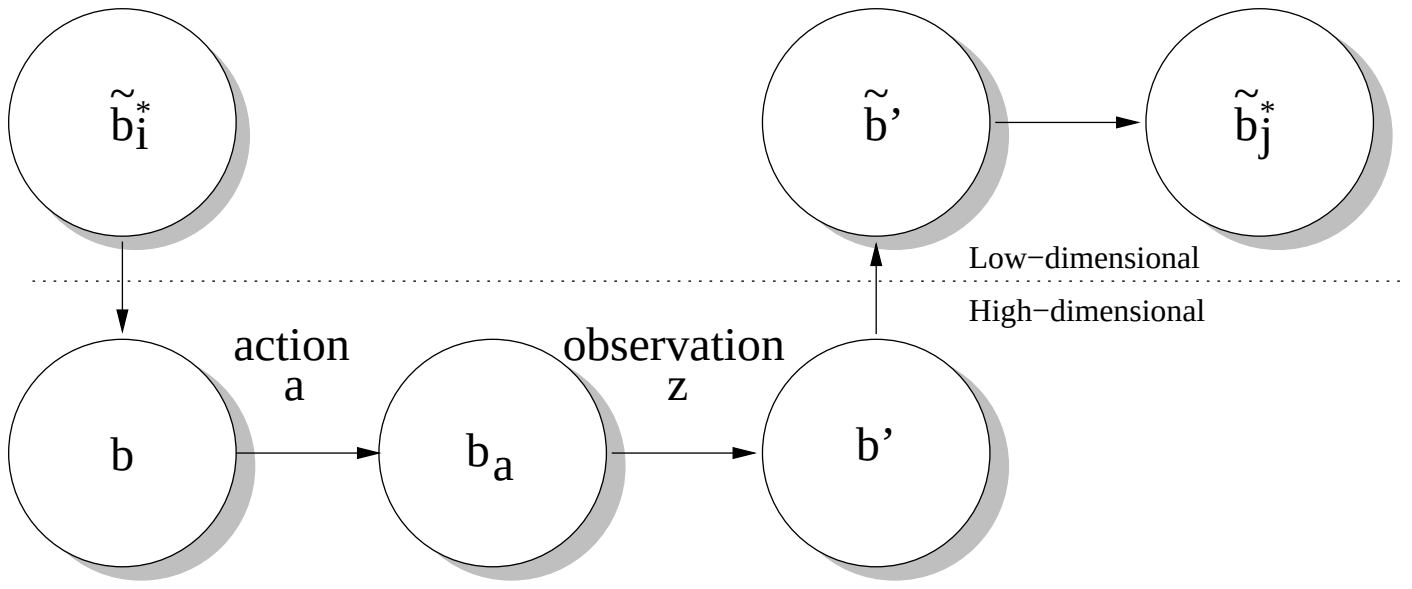

**Figure 14**: The process of computing a single transition probability.

For each transition $\tilde{b}_i^* \to b \to b_a \to b' \to \tilde{b}' \to \tilde{b}_j^*$ we can assign a probability

$$p(z, j|i, a) = p(z|b_a)\, w(\tilde{b}_j^*, \tilde{b}') = w(\tilde{b}_j^*, \tilde{b}') \sum_{l=1}^{|\mathcal{S}|} p(z|s_l) b_a(s_l) \tag{36}$$

The total transition probability $\tilde{T}^*(\tilde{b}_i^*, a, \tilde{b}_j^*)$ is the sum, over all observations $z$, of $p(z, j|i, a)$. Step 3 in Table 2 performs this computation, but shares work between the computation of $\tilde{T}^*(\tilde{b}_i^*, a, \tilde{b}_j^*)$ for different posterior beliefs $\tilde{b}_j$ which are reachable from the same prior belief $\tilde{b}_i$ under action $a$.

### Computing the Value Function

With the reward and transition functions computed in the previous sections, we can use value iteration to compute the value function for our belief space MDP. The full algorithm is given in Table 2.

## 7. Solving Large POMDPs

In this section, we present the application of our algorithm to finding policies for large POMDPs.

### Toy problem

We first tested the E-PCA belief features using a regular grid representation on a version of the toy problem described earlier. To ensure that we only needed a small set of belief samples $\tilde{b}_i^*$, we made the goal region larger. We also used a coarser discretization of the underlying state space (40 states instead of 200) to allow us to compute the low-dimensional model more quickly.

Figure 15 shows a comparison of the policies from the different algorithms. The E-PCA does approximately twice as well as the Maximum-Likelihood heuristic; this heuristic guesses its corridor, and is correct only about half the time. The "AMDP Heuristic" algorithm is the Augmented MDP algorithm reported by Roy and Thrun (1999). This controller attempts

1. Generate the discrete low-dimensional belief space $\tilde{B}^*$ using E-PCA (cf. Table 1)
2. Compute the low-dimensional reward function $\tilde{R}^*$:

   For each $\tilde{b}^* \in \tilde{B}^*, a \in \mathcal{A}$

   (a) Recover $b$ from $\tilde{b}^*$

   (b) Compute $\tilde{R}^*(\tilde{b}, a) = \sum_{i=1}^{|\mathcal{S}|} R(s_i, a) b(s_i)$.
3. Compute the low-dimensional transition function $\tilde{T}^*$:

   For each $\tilde{b}_i^* \in \tilde{B}^*, a \in \mathcal{A}$

   (a) For each $\tilde{b}_j^* : \tilde{T}^*(\tilde{b}_i^*, a, \tilde{b}_j^*) = 0$

   (b) Recover $b_i$ from $\tilde{b}_i^*$

   (c) For each observation $z$

   (d) Compute $b_j$ from the Bayes' filter equation (33) and $b$.

   (e) Compute $\tilde{b}'$ from $b_j$ by iterating equation (26).

   (f) For each $\tilde{b}_j^*$ with $w(\tilde{b}_j^*, \tilde{b}') > 0$

   (g) Add $p(z, j|i, a)$ from equation (36) to $\tilde{T}^*(\tilde{b}_i^*, a, \tilde{b}_j^*)$
4. Compute the value function for $\tilde{B}^*$

   (a) t = 0

   (b) For each $\tilde{b}_i^* \in \tilde{B}^* : V^0(\tilde{b}_i^*) = 0$

   (c) do

   (d) change = 0

   (e) For each $\tilde{b}_i^* \in \tilde{B}^*$:

   $$V^t(\tilde{b}_i^*) = \max_a \left( \tilde{R}^*(\tilde{b}_i^*, a) + \gamma \sum_{j=1}^{|\tilde{B}^*|} \tilde{T}^*(\tilde{b}_i^*, a, \tilde{b}_j^*) \cdot V^{t-1}(\tilde{b}_j^*) \right)$$

   $$\text{change} = \text{change} + V^t(\tilde{b}_i^*) - V^{t-1}(\tilde{b}_i^*)$$

   (f) while change $> 0$

**Table 2**: Value Iteration for an E-PCA POMDP

to find the policy that will result in the lowest-entropy belief in reaching the goal. This controller does very poorly because it is unable to distinguish between a unimodal belief that knows which corridor it is in but not its position within the corridor, and a bimodal belief that knows its position but not which corridor. The results in Figure 15 are averaged over 10,000 trials.

It should be noted that this problem is sufficiently small that conventional PCA fares reasonably well. In the next sections, we will see problems where the PCA representation does poorly compared to E-PCA.

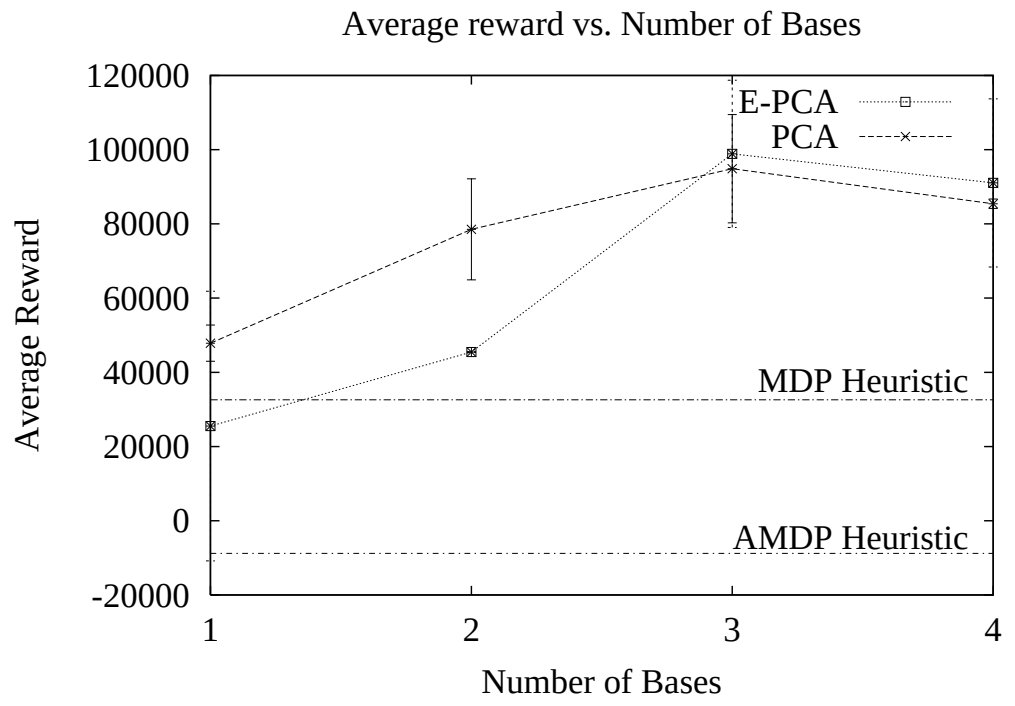

**Figure 15**: A comparison of policy performance using different numbers of bases, for 10,000 trials, with regular grid discretization. Policy performance was given by total reward accumulated over trials.

### Robot Navigation

We tested the E-PCA POMDP algorithm on simulated robot navigation problems in two example environments, the Wean Hall corridor shown in Figure 16 and the Longwood retirement facility shown in Figure 1(b). The model parameters are given by robot navigation models (see Fox et al., 1999).

We evaluated the policy for the relatively simple problem depicted in Figure 16. We set the robot's initial belief such that it may have been at one of two locations in the corridor, with the objective to get to within $0.1m$ of the goal state (each grid cell is $0.2m \times 0.2m$). The controller received a reward of $+1000$ for arriving at the goal state and taking an `at_goal` action; a reward of $-1000$ was given for (incorrectly) taking this action at a non-goal state. There was a reward of $-1$ for each motion. The states used for planning in this example were the 500 states along the corridor, and the actions were forward and backward motion.

Figure 16 shows a sample robot trajectory using the E-PCA policy and 5 basis functions. Notice that the robot drives past the goal to the lab door in order to verify its orientation before returning to the goal; the robot does not know its true position, and cannot know that it is in fact passing the goal. If the robot had started at the other end of the corridor, its orientation would have become apparent on its way to the goal.

Figure 17 shows the average policy performance for three different techniques. The Maximum-Likelihood heuristic could not distinguish orientations, and therefore approximately 50% of the time declared the goal in the wrong place. We also evaluated a policy learned using the best 5 bases from conventional PCA. This policy performed substantially better than the maximum-likelihood heuristic in that the controller did not incorrectly declare that the robot had arrived at the goal. However, this representation could not detect when the robot *was* at the goal, and also chose sub-optimal (with respect to the E-PCA policy) motion actions regularly. The E-PCA outperformed the other techniques in this example because it was able to model its belief accurately, in contrast to the result in Figure 15 where PCA had sufficient representation to perform as well or better than E-PCA.

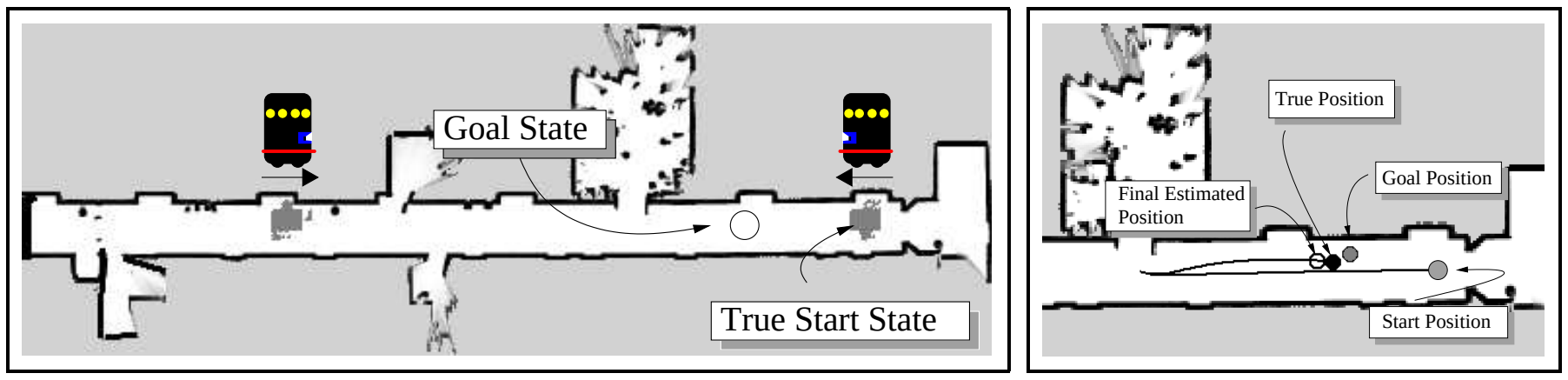

**Figure 16**: An example robot trajectory, using the policy learned using 5 basis functions. On the left are the start conditions and the goal. On the right is the robot trajectory. Notice that the robot drives past the goal to the lab door to localize itself, before returning to the goal.

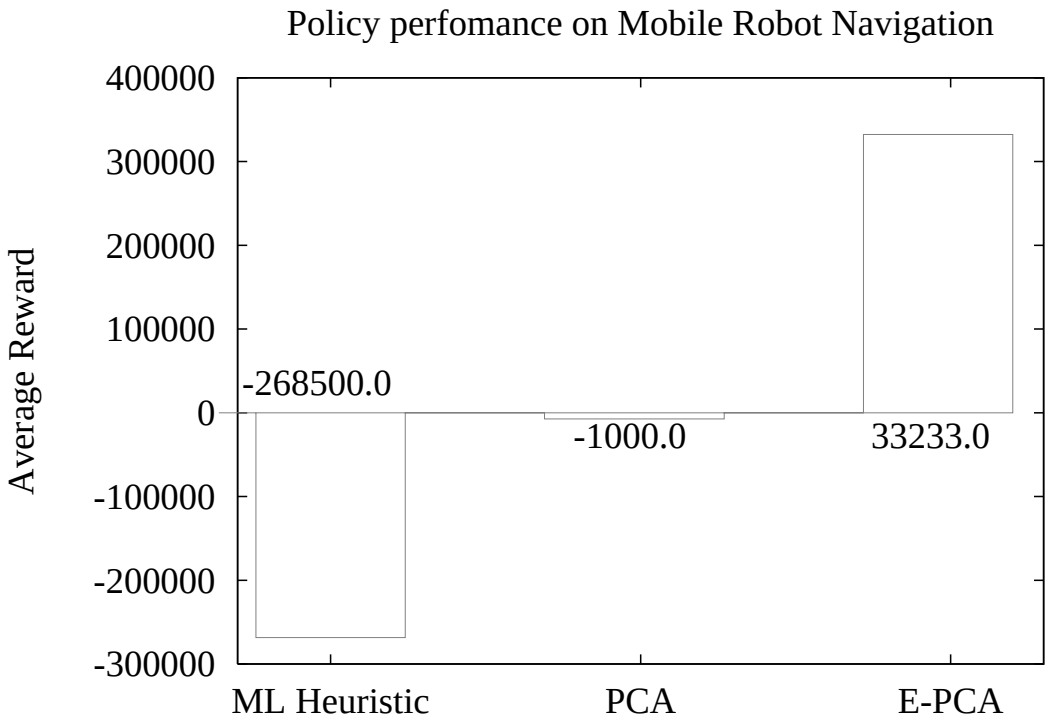

**Figure 17**: A comparison of policy performance using E-PCA, conventional PCA and the Maximum Likelihood heuristic, for 1,000 trials.

Figure 18(a) shows a second example of navigation in simulation. Notice that the initial belief for this problem is bi-modal; a good policy will take actions to disambiguate the modes before proceeding to the goal. Using a sample set of 500 beliefs, we computed the low-dimensional belief space $\tilde{B}$. Figure 18(b) shows the average KL divergence between the original and reconstructed beliefs. The improvement in the KL divergence error measure slowed down substantially around 6 bases; we therefore used 6 bases to represent the belief space.

Figure 18(c) shows an example execution of the policy computed using the E-PCA. The reward parameters were the same as in the previous navigation example. The robot parameters were maximum laser range of $2m$, and high motion model variance. The first action the policy chose was to turn the robot around and move it closer to the nearest wall. This had the effect of eliminating the second distribution mode on the right. The robot then followed essentially a “coastal” trajectory up the left-hand wall in order to stay localized, although the uncertainty in the $y$ direction became relatively pronounced. We see that as the uncertainty eventually resolved itself at the top of the image, the robot moved to the goal.

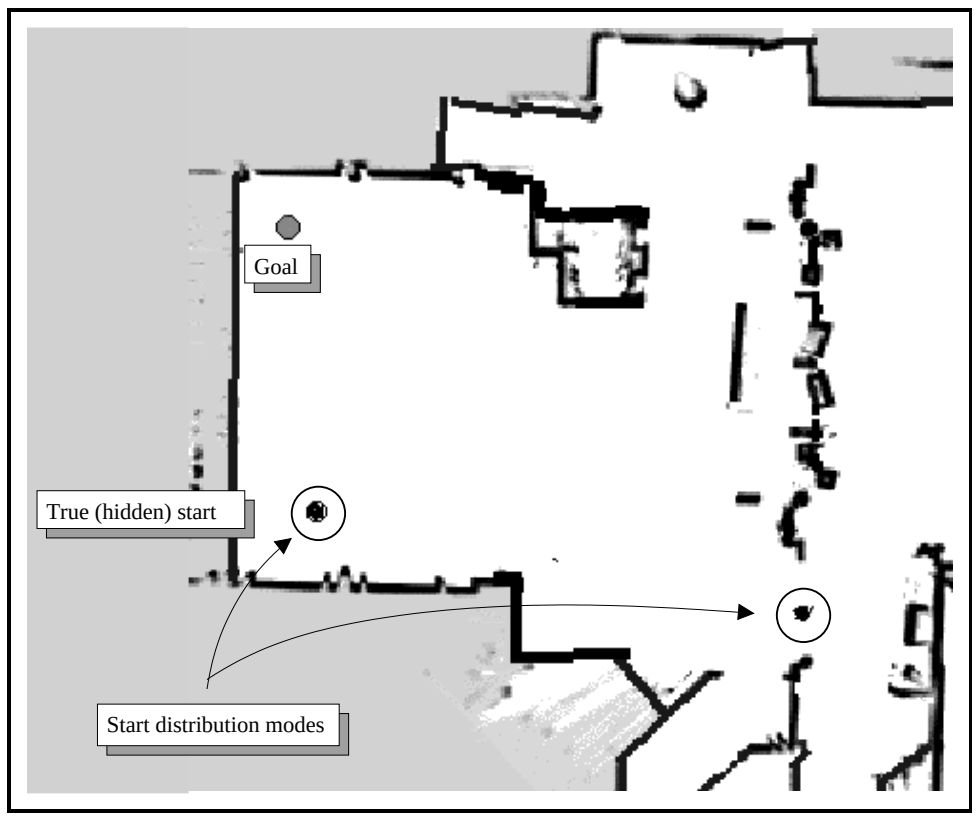

(a) Initial Distribution

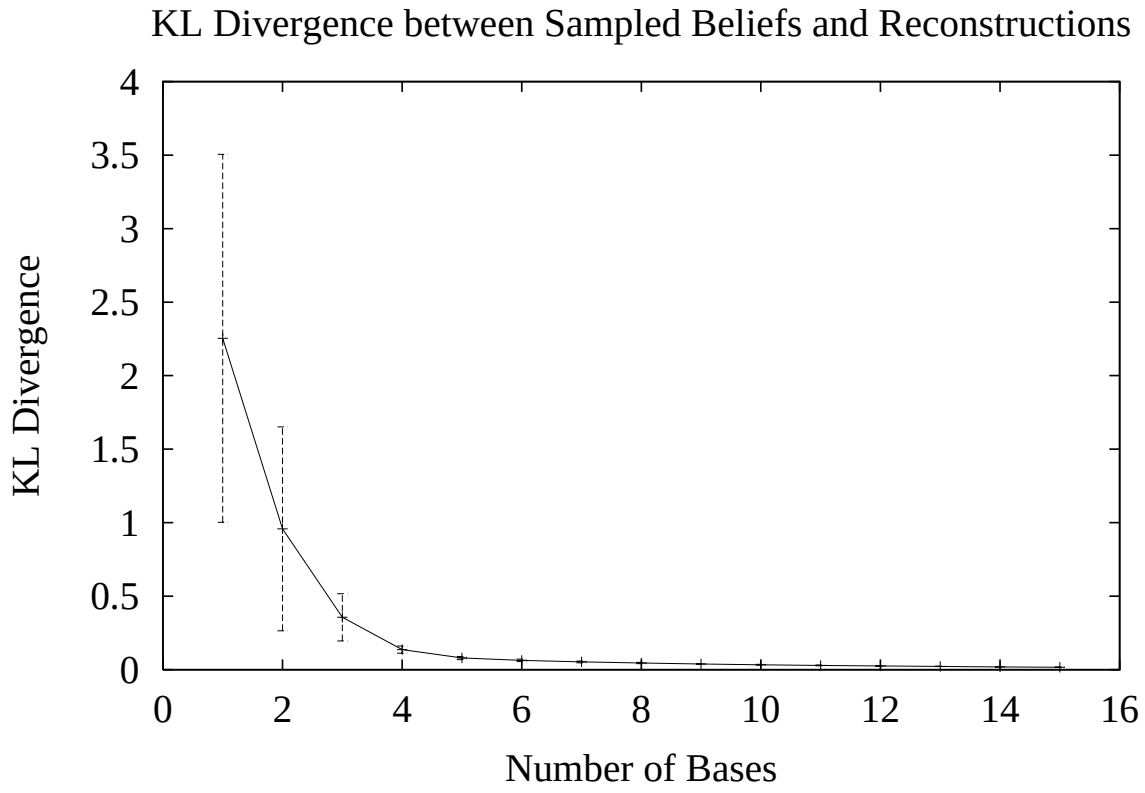

(b) Reconstruction Performance

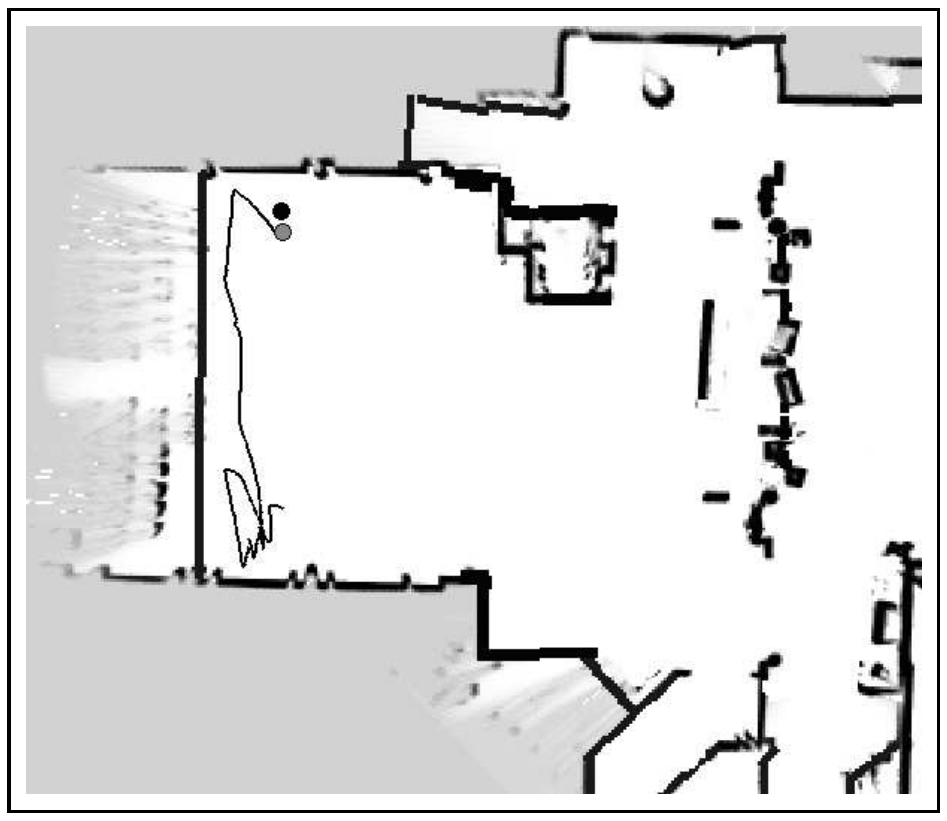

(c) The Complete Trajectory

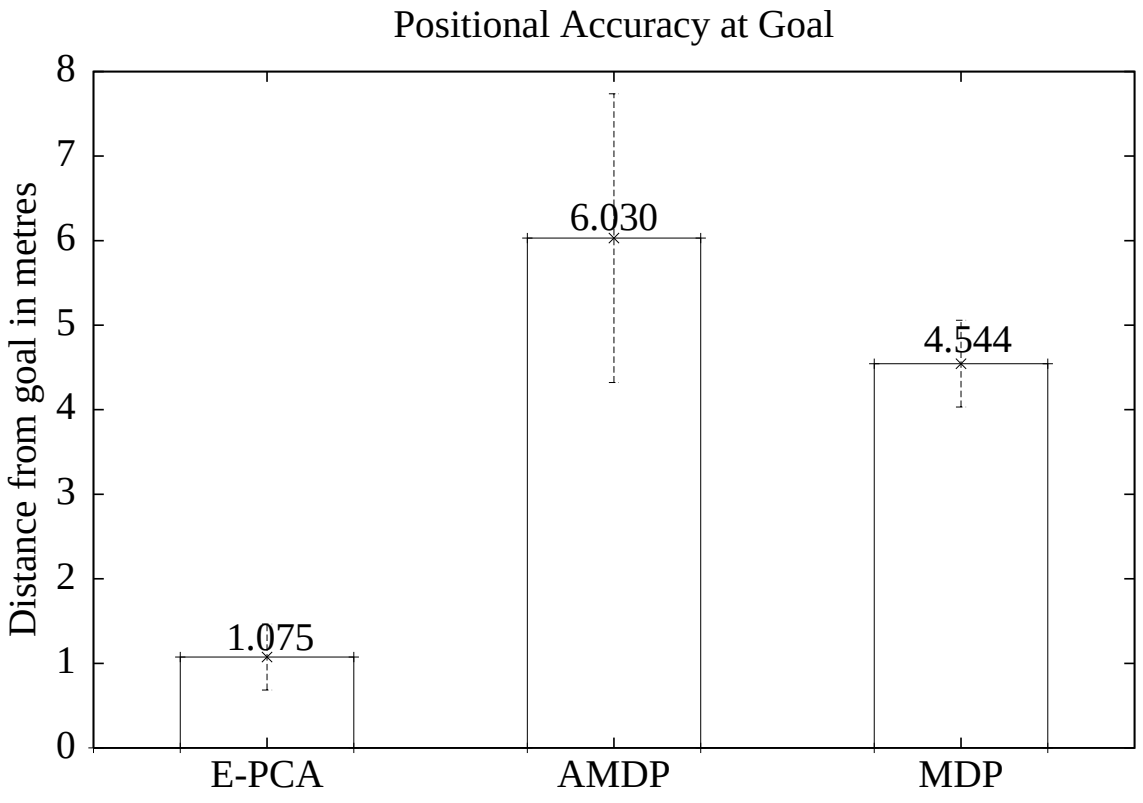

(d) Policy Performance

**Figure 18**: (a) The sample navigation problem in Longwood, cf. Figure 2. This problem involves multi-modal distributions. (c) The average KL divergence between the sample beliefs and their reconstructions against the number of bases used, for 500 samples beliefs for a navigating mobile robot in this environment. (d) A comparison of policy performance using E-PCA, conventional MDP and the AMDP heuristic.

It is interesting to note that this policy contains a similar "coastal" attribute as some heuristic policies (e.g., the Entropy heuristic and the AMDP, Cassandra, Kaelbling, & Kurien, 1996; Roy & Thrun, 1999). However, unlike these heuristics, the E-PCA representation was able to reach the goal more accurately (that is, get closer to the goal). This representation was successful because it was able more accurately to represent the beliefs and the effects of actions on the beliefs.

### Finding People

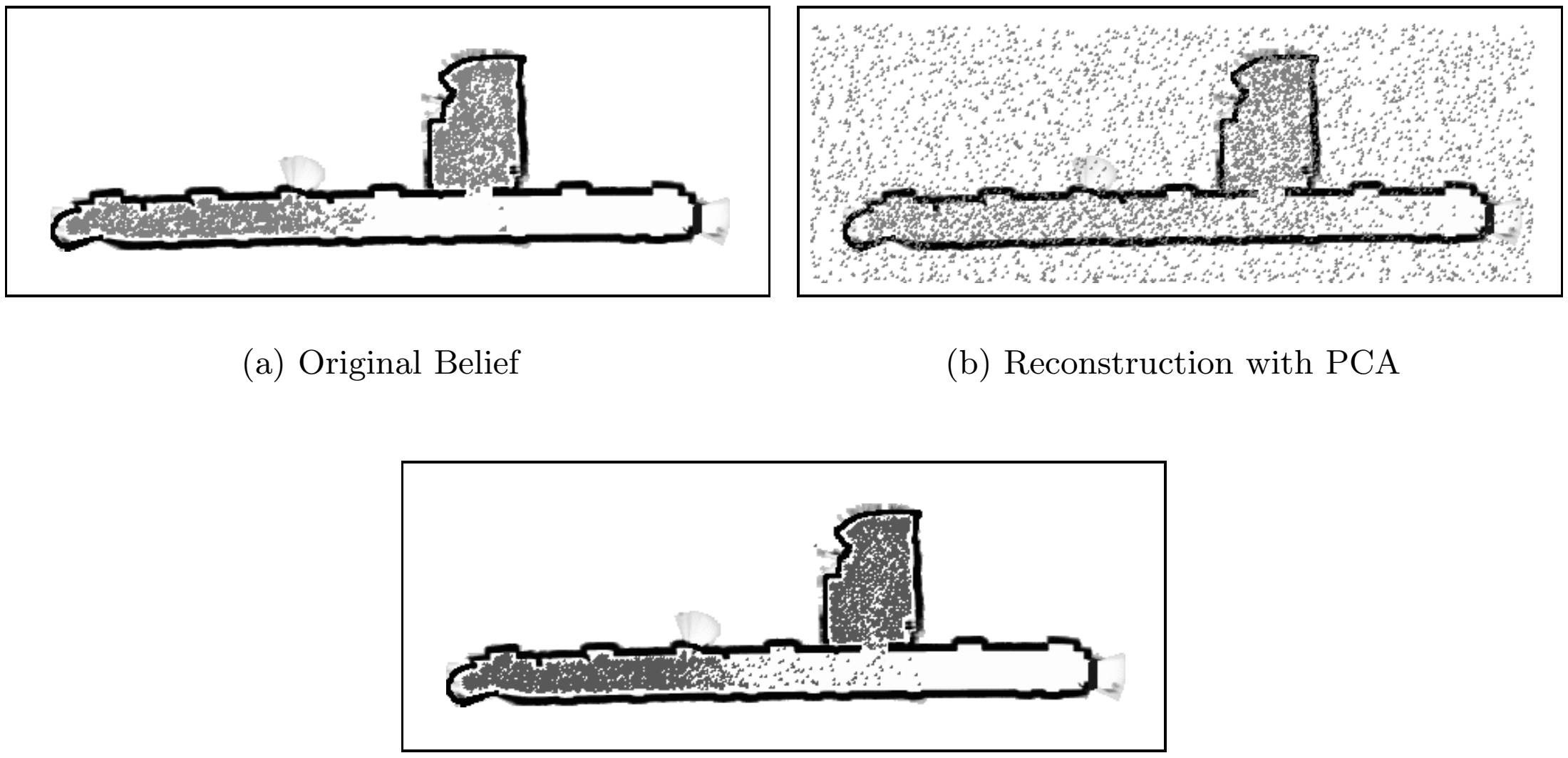

(a) Original Belief

(b) Reconstruction with PCA

(c) Reconstruction with E-PCA

**Figure 19**: The performance of PCA and E-PCA on a sample belief. The map is $238 \times 85$ grid cells, at a $0.2m$ resolution. (a) A sample belief. (b) The PCA reconstruction, using 40 bases. (c) The E-PCA reconstruction, using 6 bases.

In addition to the synthetic problem and the robot navigation problems described in the previous sections, we also tested our algorithm on a more complicated POMDP problem, that of finding a person or object moving around in an environment. This problem is motivated from the Nursebot domain, where residents experiencing cognitive decline can sometimes become disoriented and start to wander. In order to make better use of the health-care providers' time, we would like to use a robot such as Pearl (Figure 1a) to find the residents quickly. We assume that the person is not adversarial.

The state space in this problem is much larger than the previous robot navigation problems: it is the cross-product of the person's position and the robot's position. However, we assume for simplicity that the robot's position is known, and therefore the belief distribution is only over the person's position. The transitions of the person state feature are modelled by Brownian motion with a fixed, known velocity, which models the person's motion as random, independent of the robot position. (If the person was moving to avoid being "captured" by the robot, a different transition model would be required.) We assume that the position of the person is unobservable until the robot is close enough to see the person (when the robot has line-of-sight to the person, up to some maximum range, usually 3 metres); the observation model has 1% false negatives and no false positives. The reward function is maximal when the person and the robot are in the same location.

Figure 19(a) shows an example probability distribution that can occur in this problem (not shown is the robot's position). The grey dots are particles drawn from the distribution of where the person could be in the environment. The distribution is initially uniform over the reachable areas (inside the black walls). After the robot receives sensor data, the probability mass is extinguished within the sensor range of the robot. As the robot moves around, more of the probability mass is extinguished, focusing the distribution on the remaining places the person can be. However, the probability distribution starts to recover mass in places the robot visits but then leaves. In the particle filter, this can be visualized as particles leaking into areas that were previously emptied out.

We collected a set of 500 belief samples using a heuristic controller given by driving the robot to the maximum likelihood location of the person, and used E-PCA to find a good low-dimensional representation of the beliefs. Figure 19(b) shows the reconstruction of the example belief in Figure 19(a), using conventional PCA and 40 bases. This figure should reinforce the idea that PCA performs poorly at representing probability distributions. Figure 19(c) shows the reconstruction using E-PCA and 6 bases, which is a qualitatively better representation of the original belief.

Recall from section 6 that we use a function approximator for representing the value function. In the preceding examples we used a regular grid over the low-dimensional surface which performed well for finding good policies. However, the problem of finding people empirically requires a finer resolution representation than would be computationally tractable with a regular grid. We therefore turn to a different function approximator, the 1-nearest-neighbour variable resolution representation. We add new low-dimensional belief states to the model by periodically re-evaluating the model at each grid cell, and splitting the grid-cell into smaller discrete cells where a statistic predicted from the model disagrees with the statistic computed from experience. A number of different statistics have been suggested for testing the model against data from the real world (Munos & Moore, 1999), such as reduction in reward variance, or value function disagreement. We have opted instead for a simpler criterion of transition probability disagreement. We examine the policy computed using a fixed representation, and also the policy computed using an incrementally refined representation. Note that we have not fully explored the effect of different variable resolution representations of the value function, e.g., using $k$-nearest-neighbour interpolations such as described by Hauskrecht (2000). These experiments are beyond the scope of this paper, as our focus is on the utility of the E-PCA decomposition. No variable resolution representation of a value function has been shown to scale effectively beyond a few tens of dimensions at best (Munos & Moore, 2002).

This problem shares many attributes with the robot navigation problem, but we see in Figure 19 and figures 20 and 21 that this problem generates spatial distributions of higher complexity. It is somewhat surprising that E-PCA is able to find a good representation of these beliefs using only 6 bases, and indeed the average KL divergence is generally higher than for the robot navigation task. Regardless, we are able to find good controllers, and this is an example of a problem where PCA performs very poorly even with a large number of bases.

Figure 20 shows an example trajectory from the heuristic control strategy, driving the robot to the maximum likelihood location of the person at each time step. The open circle is the robot position, starting at the far right. The solid black circle is the position of the person, which is unobservable by the robot until within a $3m$ range. The person starts in the room above the corridor (a), and then moves down into the corridor once the robot has moved to the far end of the corridor (b). As the robot returns to search inside the room (c) and (d), the person moves unobserved into the previously searched corridor (e). Although we have deliberately chosen an example where the heuristic performs poorly, the person is not following an unlikely or adversarial trajectory: at all times the solid black circle remains in regions of high probability. The robot's belief accurately reflects the possibility that the person will slip past, but the heuristic control algorithm has no way to take this possibility into account.

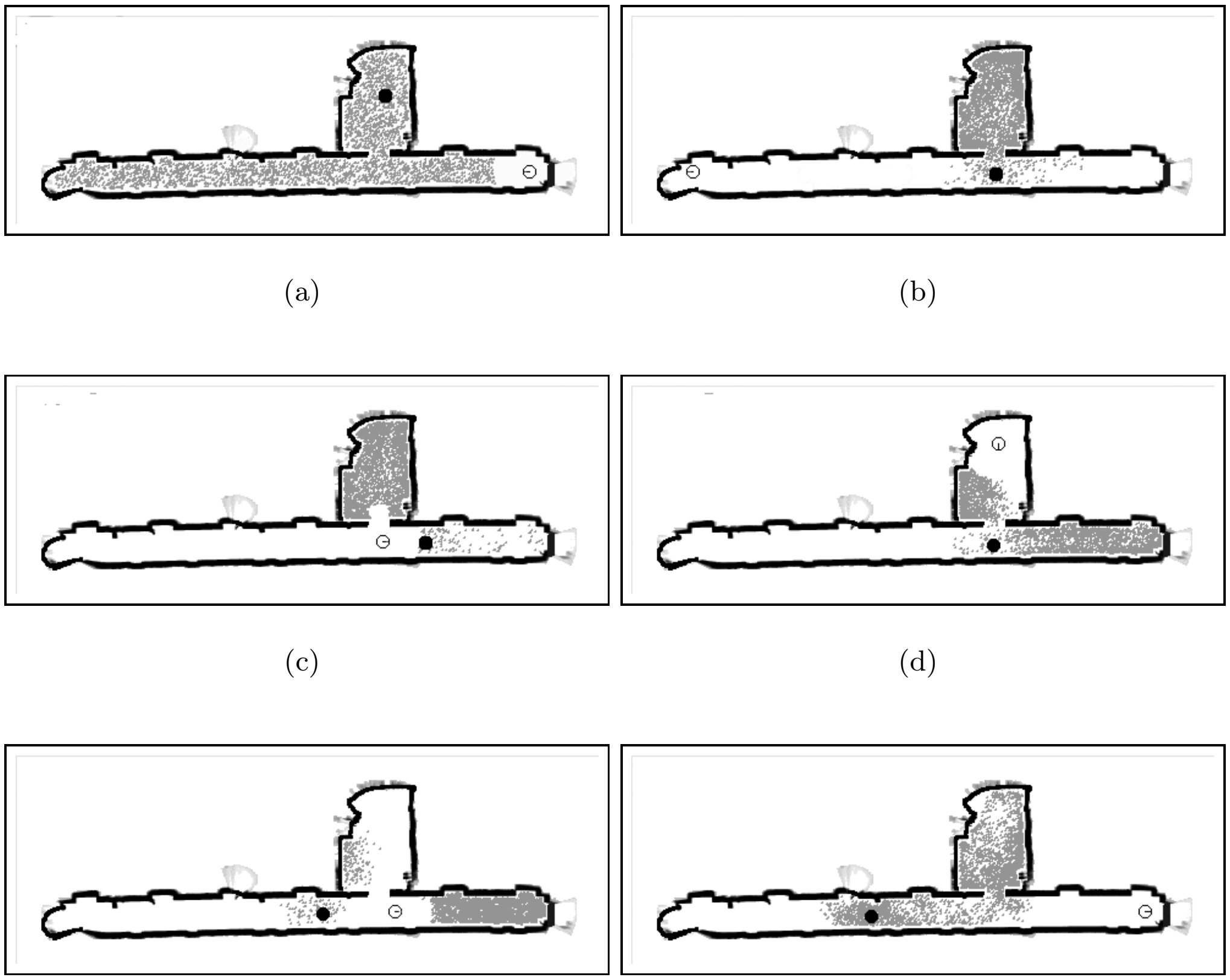

(a) (b)

(c) (d)

(e) (f)

**Figure 20**: An example of a suboptimal person finding policy. The grey particles are drawn from the distribution of where the person might be, initially uniformly distributed in (a). The black dot is the true (unobservable) position of the person. The open circle is the observable position of the robot. Through the robot's poor action selection, the person is able to escape into previously explored areas.

Using the policy found for the low-dimensional belief space as described in previous sections, we are able to find a much better controller. A sample trajectory for this controller is shown in Figure 21. The robot travels from the right-most position in the corridor (a) to only part-way down the corridor (b), and then returns to explore the room (c) and (d). In this example, the person's starting position was different from the one given in the previous example—the E-PCA policy would find the person at this point, starting from the same initial conditions as the previous example). After exploring the room and eliminating the possibility that the person is inside the room (e), the policy has reduced the possible locations of the person down to the left-hand end of the corridor, and is able to find the person reliably at that location.

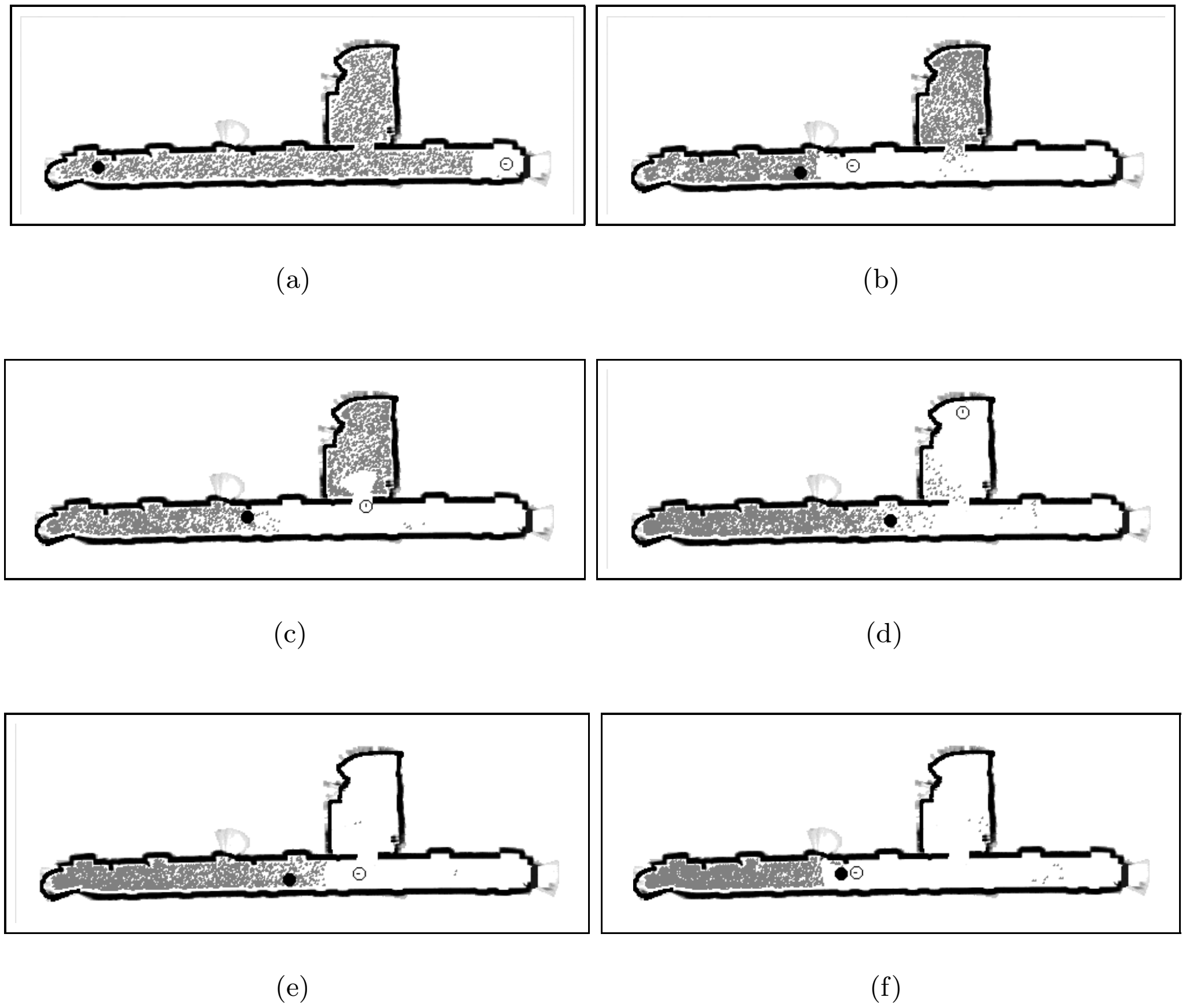

Figure 21: The policy computed using the E-PCA representation. The initial conditions in panel (a) are the same as in Figure 20. Notice that, unlike the previous figure, this strategy ensures that the probability mass is located in one place, allowing the robot to find the person with significantly higher probability.

Note that figures 20 and 21 have the target person in the worst-case start position for each planner. If the person were in the same start position in Figure 21 as in Figure 20, the policy would have found the person by panel (d). Similarly, if the person had started at the end of corridor as in Figure 21, the policy shown in Figure 20 would have found the person by panel (b).

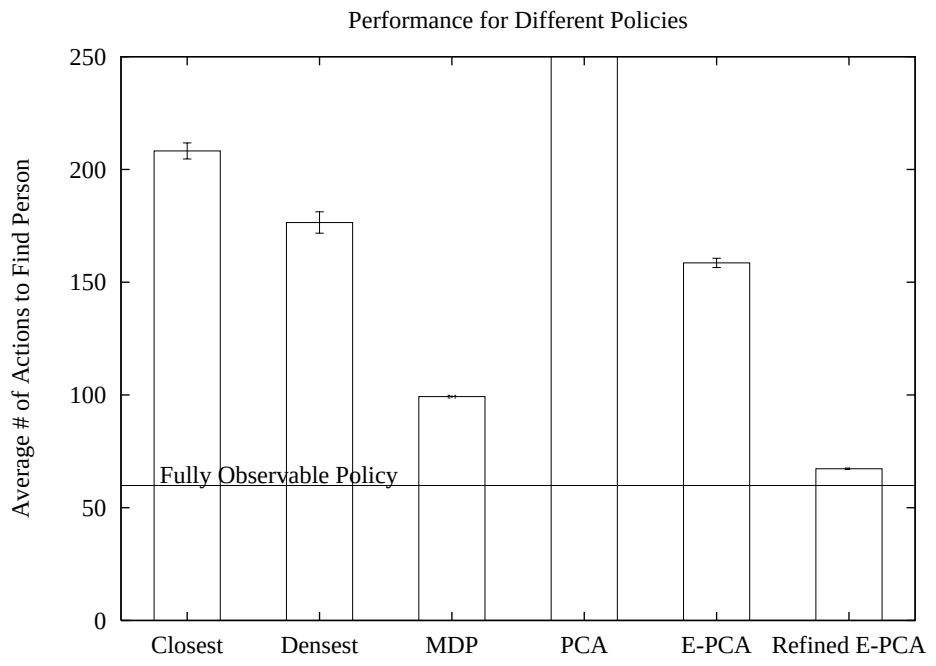

**Figure 22**: A comparison of 6 policies for person finding in a simple environment. The baseline is the fully-observable, i.e., cheating, solution (the solid line). The E-PCA policy is for a fixed (variable resolution) discretization. The Refined E-PCA is for a discretization where additional belief samples have been added. The PCA policy was approximately 6 times worse than the best E-PCA policy.

Figure 22 shows a quantitative comparison of the performance of the E-PCA against a number of other heuristic controllers in simulation, comparing the average time to find the person for these different controllers. The solid line depicts the baseline performance, using a controller that has access to the true state of the person at all times (i.e., a fully observable lower bound on the best possible performance). The travel time in this case is solely a function of the distance to the person; no searching is necessary or performed. Of course, this is not a realizable controller in reality. The other controllers are:

- Closest: The robot is driven to the nearest state of non-zero probability.
- Densest: The robot is driven to the location from which the most probability mass is visible.
- MDP: The robot is driven to the maximum-likelihood state.
- PCA: A controller found using the PCA representation and a fixed discretization of the low-dimensional surface.
- E-PCA: The E-PCA controller using a fixed discretization of the low-dimensional surface to compute the value function.
- Refined E-PCA: The E-PCA controller using an incrementally refined variable resolution discretization of the surface for computing the value function.

The performance of the best E-PCA controller is surprisingly close to the theoretical best performance, in terms of time to find the person, but this result also demonstrates the need for careful choice of discretization of the belief space for computing the value function. The initial variable resolution representation proved to be a poor function approximator, however, using the iteratively-refined variable resolution discretization, we are able to improve the performance substantially. The controller using the conventional PCA representation case was computed over a fixed discretization of the low-dimensional representation using 40 bases and 500 grid points. The quality of belief representation under PCA was so poor we did not investigate more complex policy approximators.

## 8. Discussion

These experiments demonstrate that the E-PCA algorithm can scale to finding low-dimensional surfaces embedded in very high-dimensional spaces.

### Time Complexity

The algorithm is iterative and therefore no simple expression for the total running time is available. For a data set of $|B|$ samples of dimensionality $n$, computing a surface of size $l$, each iteration of the algorithm is $O(|B|nl^2 + |B|l^3 + nl^3)$. Each step of the Newton's algorithm is dominated by a set of matrix multiplies and the final step of inverting an $l \times l$ matrix, which is $O(l^3)$. The $U$ step consists of $|B|$ iterations, where each iteration has $O(nl)$ multiplies and the $O(l^3)$ inversion. The $V$ step consists of $n$ iterations, where each iteration has $O(|B|l)$ multiplies and the $O(l^3)$ inversion, leading to the total complexity given above.

Figure 23 shows the time to compute the E-PCA bases for 500 sample beliefs, for 20,230 states. This implementation used Java 1.4.0 and Colt 1.0.2, on a 1 GHz Athlon CPU with 900M of RAM. Also shown are the computation times for conventional PCA decomposition. For small state space problems, the E-PCA decomposition can be faster than PCA for a small number of bases, if the implementation of PCA always computes the full decomposition ($l = n$, where $l$ is the reduced dimensionality and $n$ is the full dimensionality).

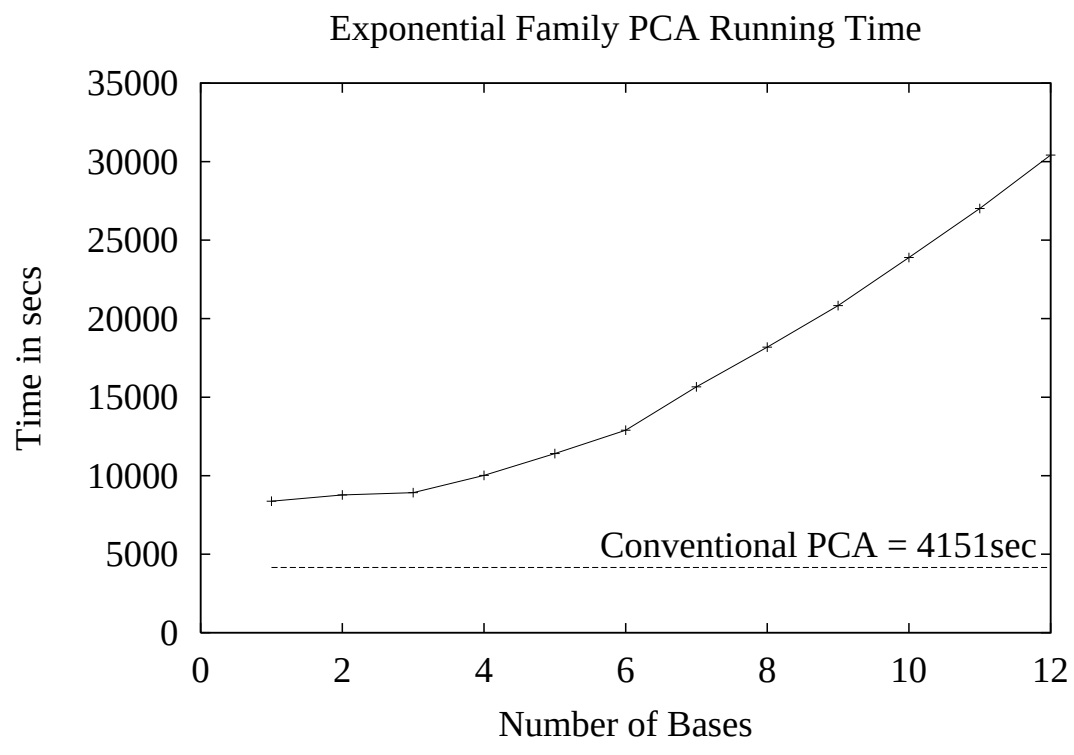

**Figure 23**: The time to compute the E-PCA representations for different discretizations of the state space.

By far the dominant term in the running time of our algorithm is the time to compute the E-PCA bases. Once the bases have been found and the low-dimensional space has been discretized, the running time required by value iteration to converge to a policy for the problems we have described was on the order of 50 to 100ms.

#### Sample Belief Collection

In all example problems we have addressed, we have used a standard sample size of 500 sample beliefs. Additionally, we have used hand-coded heuristic controllers to sample beliefs from the model. In practice, we found 500 sample beliefs collected using a semi-random controller sufficient for our example problems. However, we may be able to improve the overall performance of our algorithm on future problems by iterating between phases of building the belief space representation (i.e., collecting beliefs and generating the low-dimensional representation) and computing a good controller. Once an initial set of beliefs have been collected and used to build an initial set of bases and a corresponding policy, we can continue to evaluate the error of the representation (e.g., K-L divergence between the current belief and its low-dimensional representation). If the initial representation has been learned with too few beliefs, then the representation may over-fit the beliefs; we can detect this situation by noticing that our representation does a poor job at representing new beliefs. Validation techniques such as cross-validation may also be useful in determining when enough beliefs have been acquired.

#### Model Selection

One of the open questions we have not addressed so far is that of choosing the appropriate number of bases for our representation. Unless we have problem-specific information, such as the true number of degrees of freedom in the belief space (as in the toy example of section 3), it is difficult to identify the appropriate dimensionality of the underlying surface for control. One common approach is to examine the eigenvalues of the decomposition, which can be recovered using the orthonormalization step of the algorithm in Table 1. (This assumes our particular link function is capable of expressing the surface that our data lies on.) The eigenvalues from conventional PCA are often used to determine the appropriate dimensionality of the underlying surface; certainly the reconstruction will be lossless if we use as many bases as there are non-zero eigenvalues.

Unfortunately, recall from the description of E-PCA in section 4 that we do not generate a set of singular values, or eigenvalues. The non-linear projection introduced by the link function causes the eigenvalues of the $U$ matrix to be uninformative about the contribution of each basis to the representation. Instead of using eigenvalues to choose the appropriate surface dimensionality, we use reconstruction quality, as in Figure 11. Using reconstruction quality to estimate the appropriate dimensionality is a common choice for both PCA and other dimensionality reduction techniques (Tenenbaum, de Silva, & Langford, 2000). One alternate choice would be to evaluate the reward for policies computed for different dimensionalities and choose the most compact representation that achieves the highest reward, essentially using control error rather than reconstruction quality to determine dimensionality.

Recall from our discussion in section 2 that we are using dimensionality reduction to represent beliefs from POMDPs with a specific kind of structure. In particular, the E-PCA representation will be most useful in representing beliefs that are relatively sparse and have a small number of degrees of freedom. However, E-PCA will be unable to find good low-dimensional representations for POMDP models that do not exhibit this kind of structure – that is, if the beliefs cannot be represented as lying on low-dimensional hyperplane linked to the full belief space via the appropriate link function. One additional problem then is how to know *a priori* whether or not a specific POMDP has the appropriate structure. It is unlikely that there is a general technique that can determine the usefulness of E-PCA, but we can take advantage of model selection techniques also to determine whether or not E-PCA will find a usefully low dimensional representation for a specific POMDP. For example, if the KL divergence between a set of sample beliefs and their reconstructions is large even using a large number of bases, then the problem may not have the right structure.

## 9. Related Work

Many attempts have been made to use reachability analysis to constrain the set of beliefs for planning (Washington, 1997; Hauskrecht, 2000; Zhou & Hansen, 2001; Pineau, Gordon, & Thrun, 2003a). If the reachable set of beliefs is relatively small, then forward search to find this set is a perfectly reasonable approach. The policy computed over these beliefs is of course optimal, although it is relatively rare in real world problems to be able to enumerate the reachable beliefs. Reachability analysis has also been used with some success as a heuristic in guiding search methods, especially for focusing computation on finding function approximators (Washington, 1997; Hansen, 1998). In this approach, the problem still remains of how to compute the low-dimensional representation given the finite set of representative beliefs. Discretization of the belief space itself has been explored a number of times, both regular grid-based discretization (Lovejoy, 1991), regular variable resolution approaches (Zhou & Hansen, 2001) and non-regular variable resolution representations (Brafman, 1997; Hauskrecht, 2000). In the same vein, state abstraction (Boutilier & Poole, 1996) has been explored to take advantage of factored state spaces, and of particular interest is the algorithm of Hansen and Feng (2000) which can perform state abstraction in the absence of a prior factorization. So far, however, all of these approaches have fallen victim to the "curse of dimensionality" and have failed to scale to more than a few dozen states at most.

The value-directed POMDP compression algorithm of Poupart and Boutilier (2002) is a dimensionality-reduction technique that is closer in spirit to ours, if not in technique. This algorithm computes a low-dimensional representation of a POMDP directly from the model parameters $R$, $T$, and $O$ by finding the Krylov subspace for the reward function under belief propagation. The Krylov subspace for a vector and a matrix is the smallest subspace that contains the vector and is closed under multiplication by the matrix. For POMDPs, the authors use the smallest subspace that contains the immediate reward vector and is closed under a set of linear functions defined by the state transitions and observation model. The major advantage of this approach is that it optimizes the correct criterion: the value-directed compression will only distinguish between beliefs that have different value. The major disadvantage of this approach is that the Krylov subspace is constrained to be linear. Using our algorithm with PCA instead of E-PCA, we can realize much of the same compression as the Poupart and Boutilier (2002) method: we can take advantage of regularities in the same transition matrices $T^{a,z}$ but not in the reward function $R$. Unfortunately, as we have seen, beliefs are unlikely to lie on a low-dimensional hyperplane, and our results reported in section 3 indicate that linear compression will not scale to the size of problems we wish to address.

Possibly the most promising approaches for finding approximate value-functions are the point-based methods, which instead of optimizing the value function over the entire belief space, do so only for specific beliefs. Cheng (1988) described a method for backing up the value function at specific belief points in a procedure called "point-based dynamic programming" (PB-DP). These PB-DP steps are interleaved with standard backups as in full value iteration. Zhang and Zhang (2001) improved this method by choosing the Witness points as the backup belief points, iteratively increasing the number of such points. The essential idea is that point-based backups are significantly cheaper than full backup steps. Indeed, the algorithm described by Zhang and Zhang (2001) out-performs Hansen's exact policy-search method by an order of magnitude for small problems. However, the need for periodic backups across the full belief space still limits the applicability of these algorithms to small abstract problems.

More recently, Pineau et al. (2003a) have abandoned full value function backups in favour of only point-based backups in the "point-based value iteration" (PBVI) algorithm. By backing up *only* at discrete belief points, the backup operator is polynomial instead of exponential (as in value iteration), and, even more importantly, the complexity of the value function remains constant. PBVI uses a fundamentally different approach to finding POMDP policies, and still remains constrained by the curse of dimensionality in large state spaces. However, it has been applied successfully to problems at least an order of magnitude larger than its predecessors, and is another example of algorithms that can be used to make large POMDPs tractable.

E-PCA is not the only possible technique for non-linear dimensionality reduction; there exists a large body of work containing different techniques such as Self-Organizing Maps (Kohonen, 1982), Generative Topographic Mapping (Bishop, Svensén, & Williams, 1998), Stochastic Neighbour Embedding (Hinton & Roweis, 2003). Two of the most successful algorithms to emerge recently have are Isomap (Tenenbaum et al., 2000) and Locally Linear Embedding (Roweis & Saul, 2000). Isomap extends PCA-like methods to non-linear surfaces using geodesic distances as the distance metric between data samples, rather than Euclidean distances. Locally Linear Embedding (LLE) can be considered a local alternative to the global reduction of Isomap in that it represents each point as the weighted combination of its neighbours and operates in two phases: computing the weights of the $k$ nearest neighbours for each high-dimensional point, and then reconstructing the data in the low-dimensional co-ordinate frame from the weights. However, these algorithms do not contain explicit models of the kind of data (e.g., probability distributions) that they are attempting model. One interesting line of research, however, may be to extend these algorithms using different loss functions in the same manner that PCA was extended to E-PCA.

## 10. Conclusion

Partially Observable Markov Decision Processes have been considered intractable for finding good controllers in real world domains. In particular, the best algorithms to date for finding an approximate value function over the full belief space have not scaled beyond a few hundred states (Pineau et al., 2003a). However, we have demonstrated that real world POMDPs can contain structured belief spaces; by finding and using this structure, we have been able to solve POMDPs an order of magnitude larger than those solved by conventional value iteration techniques. Additionally, we were able to solve different kinds of POMDPs, from a simple highly-structured synthetic problem to a robot navigation problem to a problem with a factored belief space and relatively complicated probability distributions.

The algorithm we used to find this structure is related to Principal Components Analysis with a loss function specifically chosen for representing probability distributions. The real world POMDPs we have been able to solve are characterized by sparse distributions, and the Exponential family PCA algorithm is particularly effective for compressing this data. There do exist POMDP problems which do not have this structure, and for which this dimensionality reduction technique will not work well; however, it is a question for further investigation if other, related dimensionality-reduction techniques (e.g., Isomap or Locally-Linear Embedding, Tenenbaum et al., 2000; Roweis, Saul, & Hinton, 2002) can be applied.

There are a number of interesting possibilities for extending this algorithm in order to improve its efficiency or increase the domain of applicability. The loss function that we chose for dimensionality reduction was based on reconstruction error, as in

$$L(B, U, \tilde{B}) = e^{(U\tilde{B})} - B \circ U\tilde{B}, \tag{37}$$

(cf. equation 8). Minimizing the reconstruction error should allow near-optimal policies to be learned. However, we would ideally like to find the most compact representation that minimizes control errors. This could possibly be better approximated by taking advantage of transition probability structure. For example, dimensionality reduction that minimizes prediction errors would correspond to the loss function:

$$L(B, U, \tilde{B}, T) = e^{(U\tilde{b})} - B \circ U\tilde{b} + \|\tilde{B}_{\cdot,2\ldots n} - T\tilde{B}_{\cdot,1\ldots n-1}\|^2 \tag{38}$$

where $\tilde{B}_{\cdot,1\ldots n-1}$ is the $l \times n-1$ matrix of the first $n-1$ column vectors in $\tilde{B}$, and $\tilde{B}_{\cdot,2\ldots n}$ is the $l \times n-1$ matrix of the $n-1$ column vectors in $V$ starting from the second vector. This has the effect of finding a representation that allows $\tilde{b}^{t+1}$ to be predicted from $T\tilde{b}^t$, with the caveat that the $\tilde{B}$ must be arranged all for the same action. We plan to address this issue in future work.

Another shortcoming of the approach described in this work is that it contains the assumption that all beliefs can be described using the same low-dimensional representation. However, it is relatively easy to construct an example problem which generates beliefs that lie on two distinct low-dimensional surfaces, which in the current formulation would make the apparent dimensionality of the beliefs appear much higher than a set of beliefs sampled from one surface alone.

While this work has largely been motivated by finding better representations of beliefs, it is not the only approach to solving large POMDPs. Policy search methods (Meuleau,

Peshkin, Kim, & Kaelbling, 1999) and hierarchical methods (Pineau, Gordon, & Thrun, 2003b) have also been able to solve large POMDPs. It is interesting to note that controllers based on the E-PCA representations are often essentially independent of policy complexity but strongly dependent on belief complexity, whereas the policy search and hierarchical methods are strongly dependent on policy complexity but largely independent of belief space complexity. It seems likely that progress in solving large POMDPs in general will lie in a combination of both approaches.

The E-PCA algorithm finds a low-dimensional representation $\tilde{B}$ of the full belief space $B$ from sampled data. We demonstrated that the reliance on sampled data is not an obstacle for some real world problems. Furthermore, using only sampled beliefs could be an asset for large problems where generating and tracking beliefs can be considerably easier than planning. It may however be preferable to try to compute a low-dimensional representation directly from the model parameters. Poupart and Boutilier (2002) use the notion of a Krylov subspace to do this. The subspace computed by their algorithm may correspond exactly with a conventional PCA and we have seen instances where PCA does a poor job of finding low-dimensional representations. The most likely explanation is that real-world beliefs do not lie on low-dimensional planes for most problems, but instead on curved surfaces. An extremely useful algorithm would be one that finds a subset of belief space closed under the transition and observation function, but which is not constrained to find only planes.

## Acknowledgements

Thanks to Tom Mitchell, Leslie Kaelbling, Reid Simmons, Drew Bagnell, Aaron Courville, Mike Montemerlo and Joelle Pineau for useful comments and insight into this work. Nicholas Roy was funded by the National Science Foundation under ITR grant # IIS-0121426. Geoffrey Gordon was funded by AFRL contract F30602–01–C–0219, DARPA's MICA program, and by AFRL contract F30602–98–2–0137, DARPA's CoABS program.